\pdfoutput=1

\documentclass[11pt]{article}

\usepackage[]{ACL2023}

\usepackage{times}
\usepackage{latexsym}
\usepackage{float}
\usepackage{svg}
\usepackage{hyperref}
\usepackage{caption}
\usepackage{multirow}
\usepackage{makecell}
\usepackage{afterpage}
\usepackage{placeins}

\usepackage{graphicx}
\usepackage{subcaption}
\usepackage{caption}
\usepackage{placeins}
\usepackage{booktabs}
\usepackage{color}
\usepackage{lipsum}

\usepackage{amsmath}
\usepackage{amssymb}
\usepackage{mathtools}
\usepackage{amsthm}
\usepackage{enumitem}
\usepackage[normalem]{ulem}

\usepackage{xcolor}
\usepackage{soul}
\newcommand{\hlc}[2][yellow]{{%
    \colorlet{foo}{#1}%
    \sethlcolor{foo}\hl{#2}}%
}
\definecolor{myblue}{rgb}{0.651, 0.808, 0.890}
\definecolor{mygreen}{rgb}{0.698, 0.875, 0.541}
\definecolor{myorange}{rgb}{0.992, 0.749, 0.435}

\usepackage[T1]{fontenc}

\usepackage[utf8]{inputenc}

\usepackage{microtype}
\newcommand*{\affaddr}[1]{#1} 
\newcommand*{\affmark}[1][*]
{\textsuperscript{#1}}

\usepackage{inconsolata}

\title{Self-Consistent Narrative Prompts on Abductive\\
 Natural Language Inference}

\author{
Chunkit Chan\affmark[1],
Xin Liu\affmark[1],
Tszho Chan\affmark[1],
Jiayang Cheng\affmark[1],
Yangqiu Song\affmark[1],\\
\textbf{Ginny Y. Wong\affmark[2],
Simon See\affmark[2]}\\
\affaddr{\affmark[1]Department of Computer Science and Engineering, HKUST, Hong Kong SAR, China}\\
\affaddr{\affmark[2]NVIDIA AI Technology Center (NVAITC), NVIDIA, Santa Clara, USA} \\
\texttt{\{ckchancc, xliucr, zchencj, jchengaj, yqsong\}@cse.ust.hk} \\
\texttt{\{gwong, ssee\}@nvidia.com}\\
}

\begin{document}
\maketitle
\begin{abstract}
Abduction has long been seen as crucial for narrative comprehension and reasoning about everyday situations. 
The abductive natural language inference ($\alpha$NLI) task has been proposed, and this narrative text-based task aims to infer the most plausible hypothesis from the candidates given two observations. 
However, the inter-sentential coherence and the model consistency have not been well exploited in the previous works on this task. In this work, we propose a prompt tuning model \textbf{$\alpha$-PACE}\footnote{The source code is available at~\url{https://github.com/HKUST-KnowComp/Alpha-PACE}}, which takes self-consistency and inter-sentential coherence into consideration. Besides, we propose a general self-consistent framework that considers various narrative sequences (e.g., linear narrative and reverse chronology) for guiding the pre-trained language model in understanding the narrative context of input. We conduct extensive experiments and thorough ablation studies to illustrate the necessity and effectiveness of \textbf{$\alpha$-PACE}. The performance of our method shows significant improvement against extensive competitive baselines.
\end{abstract}

\section{Introduction}
\begin{figure}[t]
    \centering
    \includegraphics[width=\linewidth]{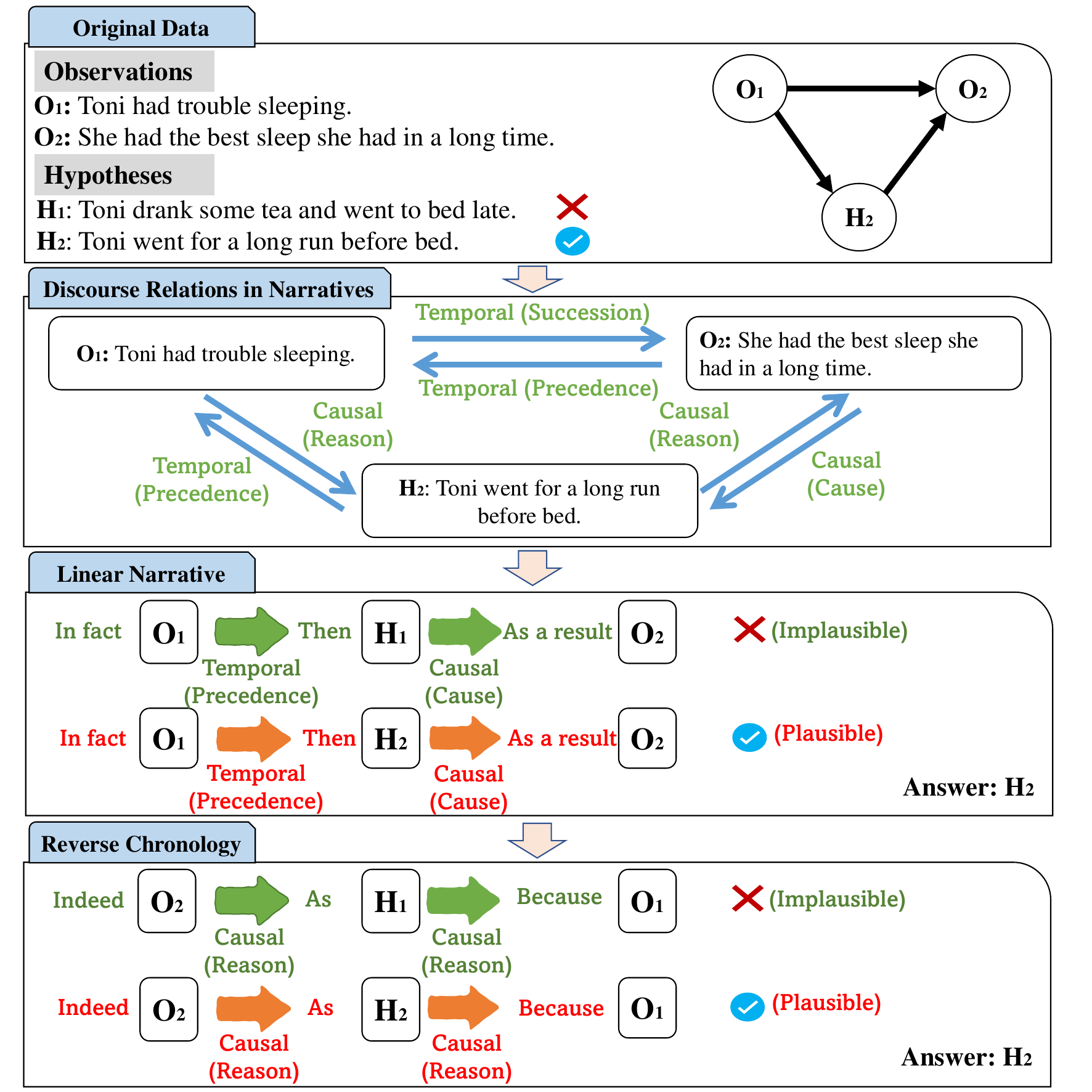}
    \vspace{-0.6cm}
    \caption{
    A data example from $\alpha$NLI and its corresponding narrative sequences, including linear narrative and reverse chronology. Two sequences explain the same narrative example seamlessly by utilizing the discourse connectives (i.e., `` in fact,'' `` then,'' `` as a result,'').}
    \label{fig:DataInstance_Introduction}
    \vspace{-0.4cm}
\end{figure}

Abductive reasoning aims to find the most plausible explanation based on incomplete observations~\cite{peirce1974collected}. Abduction has long been seen to be essential for understanding narratives~\cite{hobbs1993interpretation} and reasoning about everyday situations~\cite{andersen1973abductive}. 
\citet{bhagavatula2020abductive} investigated the language-based abduction in narrative texts and introduced the abductive natural language inference ($\alpha$NLI) benchmark, which is a multiple-choice question answering task for identifying the most likely explanation among two hypotheses based on two observations. One example is illustrated in Figure~\ref{fig:DataInstance_Introduction}, where ``$O_1$'' and ``$O_2$'' are two observations. Abductive reasoning is identifying a possible hypothesis (either $H_1$ or $H_2$) that can best explain the consequences by evaluating and comparing the plausibility of these two hypotheses. 

Traditional works on the $\alpha$NLI task focus on ranking the hypotheses among ``$H_1$'' and ``$H_2$''~\cite{zhu2020l2r2,li2021interactive} or incorporating the knowledge from various sources into pre-trained language models, such as general commonsense knowledge~\cite{Mitra2020HowAK, du2021learning} and social commonsense knowledge~\cite{paul2020social}. 
However, one crucial piece of information, i.e., the inter-sentential coherence and the consistency of the model, has yet to be investigated and explored.

These prior studies often concatenate the observations and hypotheses as the model input, ignoring the coherence between sentences and their inter-sentential relations in this narrative-based task. Narrative, as a semiotic representation of a sequence of events meaningfully connected in a temporal and causal way~\cite{ryan2007toward,onega2014narratology}, intrinsically encodes the information required for abductive reasoning that makes it logical, sensible, and coherent.
For instance, Figure~\ref{fig:DataInstance_Introduction} illustrates that the relation connected from ``$H_2$'' to ``$O_2$'' is a causal relation emphasized by a discourse connective ``as a result'' and provides the extra causal information needed for pre-trained language models (PLMs) to comprehend these observations and hypotheses in depth. Furthermore, the consistency of a model, a highly desirable characteristic for a model in natural language processing, refers to the invariant in behavior despite meaning-preserving alterations in its input. Prior research has highlighted the significance of mode consistency and revealed that language models could exhibit inconsistencies in various contexts, including conversation, explanation generation, and factual knowledge extraction~\cite{DBLP:journals/corr/abs-2001-09977, DBLP:conf/acl/CamburuSMLB20, DBLP:journals/tacl/ElazarKRRHSG21}. These inconsistencies may result in output variability and local optimality. Since the prompt tuning-based method reduces the model variability by freezing the pre-trained model without altering the representations, we propose a self-consistent prompt tuning model that considers the inter-sentential coherence in the $\alpha$NLI task.

We have noticed that \citet{DBLP:journals/corr/abs-2203-11171} proposed a self-consistent framework (i.e., sample-and-marginalize method) that focuses on the answer consistency among diverse reasoning paths.
It relies on an individual prompt to sample various outputs and perform majority voting to address the inconsistency issue that language models suffer.
However, this method may not be an optimal method for the $\alpha$NLI task as it does not take the narrative sequences into account. A narrative usually describes the sequence of events in various narrative orders, utilizing different inter-sentential relations. In particular, people can understand the same narrative context by utilizing alternative narrative sequences instead of \textit{linear narrative}, such as \textit{nonlinear narrative} and \textit{reverse chronology}. For example, Figure~\ref{fig:DataInstance_Introduction} shows that both linear narrative and reverse chronology can explain the same narrative seamlessly by employing discourse connectives. In this linguistic phenomenon, two narrative sequences with different description orders emphasize different partial information about these events, while expressing the same narrative context.
When applying machine learning for abductive reasoning, with context sequences being different, the performance of models can vary as pre-trained language models interpret the context information from diverse perspectives and extents. 

In this paper, we attempt to imitate the cognitive process of narrative understanding, 
and propose a general self-consistent framework to facilitate a PLM understanding of the narrative context based on the above linguistic phenomenon considering different narrative sequences.
For each narrative sequence, we design a specific prompt template to distinguish the difference in narrative order while still incorporating inter-sentential coherence and self-consistency.

Our contributions are summarized as follows:

\noindent 1. This work is the first to consider inter-sentential coherence and self-consistency through the prompt tuning method in the task.

\noindent 2. We propose a general self-consistent framework based on the linguistic phenomenon that allows various narrative sequences for undertaking abductive reasoning.

\noindent 3. We conduct extensive experiments and thorough ablation studies to illustrate the necessity and effectiveness of the specific prompt template and general self-consistent framework. The results support our claims and the success of our proposed model.

\begin{figure*}[t]
    \centering
    \includegraphics[width=\linewidth]{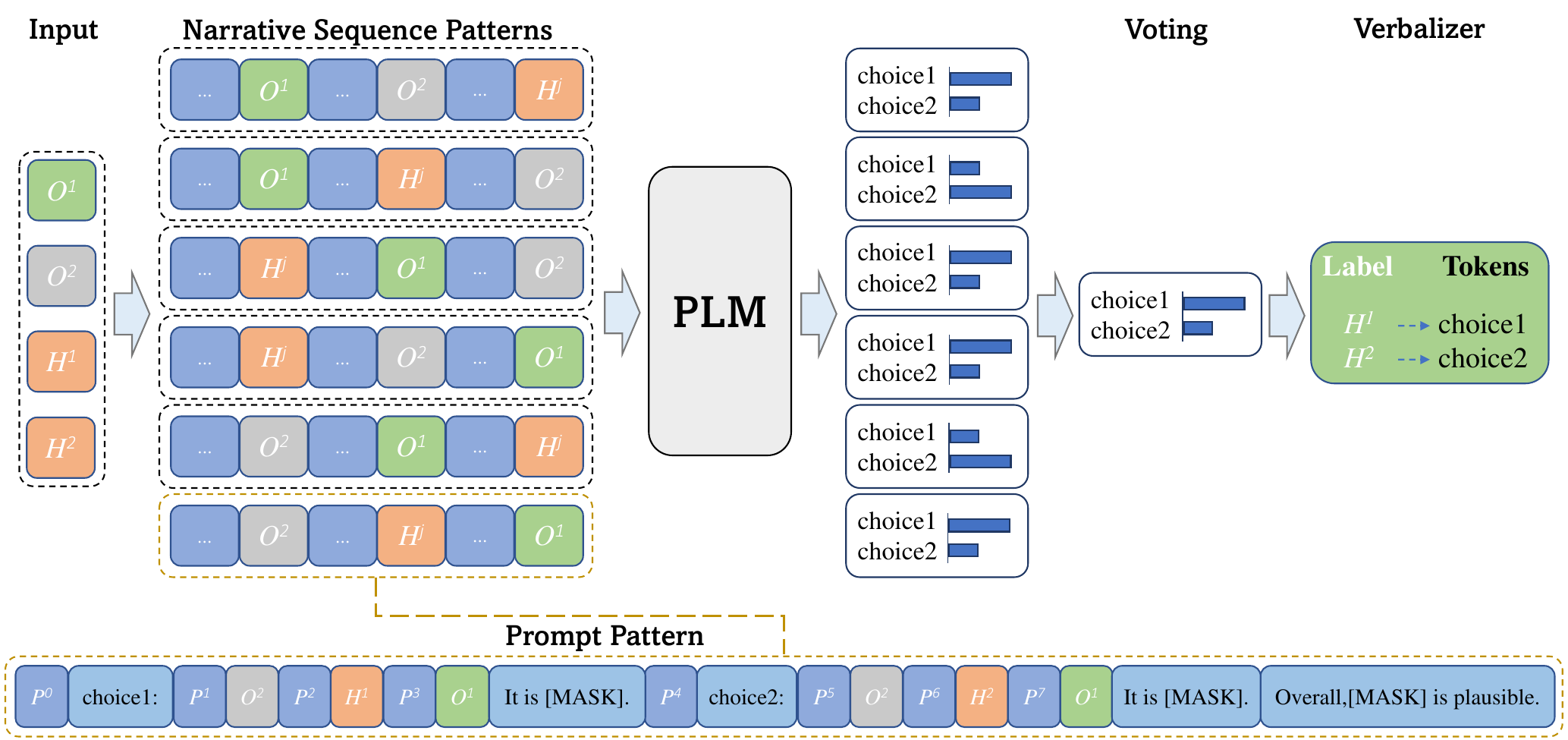}
    \vspace{-0.6 cm}
    \caption{The general self-consistent narrative prompt framework for considering varying narrative sequences. Two observations $(O^1, O^2)$ and a pair of hypotheses $(H^1, H^2)$ are permuted as six different sequence patterns, where the corresponding task-specific self-consistent prompt pattern includes two prefix prompts $P^0, P^4$, six cloze prompts $P^1, P^2, P^3, P^5, P^6, P^7$, and the manual template ``It is [MASK].'' and ``Overall, [MASK] is plausible.'' The majority voting results align to label predictions finally.}
    \label{fig:framework}
    \vspace{-0.4 cm}
\end{figure*}

\section{Related Work}
\paragraph{Abductive Reasoning} Abduction has long been thought necessary for comprehending narrative \cite{hobbs1993interpretation} and reasoning about everyday events \cite{andersen1973abductive}. Most earlier research has concentrated on formal logic-based abductive reasoning \cite{levesque1989knowledge,ng1990role,DBLP:journals/air/Paul93}. However, the rigidity of formal logic restricts its application in the field of NLP. Hence, \citet{bhagavatula2020abductive} developed a language-based abductive reasoning task to help with this, and they developed baselines that adopt the pre-trained language models (i.e., BERT \cite{devlin-etal-2019-bert}) under their probabilistic framework. To solve this task, \citet{paul2020social} proposed a multi-head knowledge attention approach to enhance RoBERTa \cite{Liu2019RoBERTaAR} by incorporating the structured social commonsense knowledge generated from COMET \cite{bosselut2019comet}. \citet{du2021learning} employed a latent variable to acquire commonsense knowledge from the event graph and enhance the pre-trained language model RoBERTa. Apart from incorporating commonsense knowledge to tackle this task, \citet{zhu2020l2r2} reformulated the $\alpha$NLI task as a ranking task using a learning-to-ranking framework to rank candidate hypotheses. \citet{li2021interactive} proposed an interactive language model that groups the correct and incorrect hypotheses instead of ranking these hypotheses and adopts joint softmax focal loss for this $\alpha$NLI task. 
However, prior works did not exploit model consistency and various narrative sequences in this task.

\paragraph{Prompt Tuning}
By relaxing the constraint that prompts token embedding to be the natural language, \citet{DBLP:conf/acl/LiL20} and \citet{hambardzumyan-etal-2021-warp} proposed combining a PLM's input token embeddings with additional continuous vectors. Some studies~\cite{DBLP:conf/emnlp/LesterAC21,qin2021learning,DBLP:conf/acl/LiL20} proposed only tuning continuous prompts, while some works \cite{han2021ptr,zhong2021factual, liu2021gpt, DBLP:conf/acl/ChanLCLSWS23} explore combining discrete prompts and continuous prompts. They tune the embedding of these additional continuous vectors, and the parameters of PLMs are frozen in their task. In our work, we also adopt this strategy, but we focus on utilizing this approach to investigate the model consistency and the narrative coherence information underlying various observations and hypotheses in this task.

\section{$\alpha$-PACE}
In order to explore the inter-sentential coherence and model consistency for the abductive natural language inference ($\alpha$NLI) task, 
we propose the \textbf{a}bductive self-consistent \textbf{P}rompt tuning model on n\textbf{A}tural language inferen\textbf{CE} task (\textbf{$\alpha$-PACE}).

\subsection{Problem Definition}
Abduction is to infer the most reasonable explanation for incomplete observations~\cite{peirce1974collected}.
In the $\alpha$NLI task~\cite{bhagavatula2020abductive}, given two observations $O_i^1$ and $O_i^2$, we choose the most plausible hypothesis among $H_i^1$ and $H_i^2$:
\\
\scalebox{0.91}{\parbox{1.10\linewidth}{
\begin{align}
    H_{i}^{*}=\arg \max _{H_{i}^{j}} P\left(H_{i}=H_{i}^{j} \mid O_{i}^{1}, O_{i}^{2}\right),
\end{align}
}}
\\
where $H_{i}^{*}$ is the most reasonable hypothesis, and $i$ indicates $i$-th instance of the dataset $\mathcal{D}=\left\{\left(x_{i},y_{i}\right)\right\}_{i=1}^{|\mathcal{D}|}$ where $x_{i}=\left\{O_{i}^{1}, O_{i}^{2}, H_{i}^{1}, H_{i}^{2}\right\}$.
We will omit the index $i$ without causing ambiguity in the following part.

\subsection{T5 Foundation Model}
T5~\cite{DBLP:journals/jmlr/RaffelSRLNMZLL20}, an encoder-decoder model, has been pre-trained on a multi-task mixture of unsupervised and supervised tasks. The unsupervised denoising training task focused on training this model to predict consecutive masked spans of tokens. For instance, the input ``She had the best sleep she had in a long time.'' was corrupted as ``She <X> the best sleep she had in a <Y>.'' The target output was ``<X> had <Y> long time </s>'' </s> is the eos\_token. The supervised pre-trained task required the model to perform a sequence-to-sequence input-output mapping with the instruction of a task prefix (e.g., ``translate German to English:'' or ``summarize:''). However, discovering the specific textual prefix token was arduous and required enormous human effort. To overcome this issue, prefix tuning~\cite{DBLP:conf/acl/LiL20} and prompt tuning~\cite{DBLP:conf/emnlp/LesterAC21} methods were proposed, which relaxed the constraint of discrete textual tokens to continuous and tunable ones.

\subsection{Self-Consistent Prompt Tuning Model}
To predict the hypothesis $H_{i}^{*}$ for each instance $x_i$, we employed a human-tailored template $\mathcal{T}(\cdot)$ transforming the data instances $x_i$ to the prompt input $\tilde{x}_i = \mathcal{T}(x_i)$, and a verbalizer V(·) is utilized to map a set of words to class labels. Figure~\ref{fig:framework} illustrates the architecture of \textbf{$\alpha$-PACE}.

\subsubsection{Task-Specific Self-Consistent Prompt}
The meticulously devised template, crafted to investigate inter-sentential coherence and self-consistency, comprises essential discrete tokens, masked tokens, and learnable continuous tokens. 

\paragraph{Inter-Sentential Coherence} For the inter-sentential coherence, we concatenate the $O_{i}^{1}$, $H_{i}^{1}$, $O_{i}^{2}$ and $O_{i}^{1}$, $H_{i}^{2}$, $O_{i}^{2}$ as two sentence sequences $S_{i}^{1}$ and $S_{i}^{2}$, instead of directly connect the $O_{i}^{1}$, $O_{i}^{2}$, $H_{i}^{1}$, and $H_{i}^{2}$ together as the model input.
These two sequences are to help PLMs easily capture the coherence information inherently in various sequences. Moreover, \citet{chatman1980story} explains that the story is the context of narrative (the what of the narrative), and the discourse is the form of narrative (the how). Specifically, the discourse is the means by which the narrative content is expressed \cite{chatman1980story,tomavsvcikova2009narrative}. Therefore, by adding discourse connectives between two events, the pre-trained language model can understand the narrative context more easily and enhance the inter-sentential coherence. Nevertheless, employing diverse discourse connectives for each data instance presents a formidable challenge. Hence, we insert the continuous tunable prompt tokens to represent the discourse connectives between each sentence (i.e.,$O_{i}^{1}$, $H_{i}^{1}$, $O_{i}^{2}$ and $O_{i}^{1}$) to learn the coherence information between these sentences. Since some discourse connectives naturally start before the first sentence (such as ``since''  and ``although''), we assign the continuous tunable prompt tokens before the first sentence of the sentence sequence. We follow \citet{DBLP:journals/corr/abs-2107-13586} to name the continuous prompts in our method. The continuous prompts are denoted as $\{\mathrm{P}^{k} \in \mathbb{R}^{p^{k} \times d} | k=0, 1, \cdots, 7\}$, where the $\mathrm{P}^0$ and $\mathrm{P}^4$ serve as the prefix prompt to learn the instruction guiding the model to perform the $\alpha$NLI task by following~\citet{DBLP:conf/emnlp/LesterAC21}. Other prompt tokens correspond to the cloze prompt between two different sentences (or before the first sentence) utilized to represent the discourse connectives to learn the coherence information between sentences. $p^{k}$ is the length of the $k$-th prompt. 

\paragraph{Self-Consistent Prompt}
For the self-consistency of model output, three [MASK] tokens are included: a [MASK] combined with the discrete token ``is plausible.'' forming the manual template ``Overall, [MASK] is plausible.'' for facilitating model inference, and each of another two [MASK] merges with the discrete token ``, it is'' to constitute ``, it is [MASK]'' append after each sentence sequence. Furthermore, the discrete tokens ``choice1:'' and ``choice2:'' are placed before sequences $S_{i}^{1}$ and $S_{i}^{2}$ respectively for splitting two sequences. These three [MASK] tokens are used for achieving the purpose of model self-consistency by ensuring three model outputs consistently. By considering the mentioned sentence sequences and the example in Figure~\ref{fig:DataInstance_Introduction}, humans are able to recognize that $H_{i}^{2}$ is more plausible, resulting from the sentence sequence $S_{i}^{1}$ is not plausible or less plausible than $S_{i}^{2}$. In this case, the pre-trained language model guided to predict ``not plausible'' for $S_{i}^{1}$, ``plausible'' for $S_{i}^{2}$, and ``choice2'' in the third mask in the learning process. Throughout this learning process, the model will learn the output consistency ability. Therefore, we introduce three masks in our model to predict the plausibility of $S_{i}^{1}$ and $S_{i}^{2}$, and the last mask for final determined labels (i.e., ``choice1'' or ``choice2'' representing the $H_{i}^{1}$ and $H_{i}^{2}$).

\begin{table}[t]
\centering
\small
\scalebox{0.8}{\begin{tabular}{c | c c c}
\toprule
\textbf{Class Label} & \textbf{First [MASK]} &  \textbf{Second [MASK]} &  \textbf{Third [MASK]}\\
\midrule
$H^{1}$ & plausible & not plausible  & choice1\\
$H^{2}$ & not plausible & plausible  & choice2\\
\bottomrule
\end{tabular}}
\vspace{-0.1in}
\caption{The label word set on $\alpha$NLI task.}
\label{tab:Label_Set}
\vspace{-0.2in}
\end{table}

\paragraph{Self-Consistent Verbalizer}
A typical verbalizer usually maps a label $y$ to a single answer token $z$ or a series of spans $z^1, z^2, \cdots $ greedily ~\cite{DBLP:conf/eacl/SchickS21,DBLP:journals/corr/abs-2107-13586}.
We extend it by mapping two class labels (i.e., $H_{i}^{1}$ and $H_{i}^{2}$ ) to three tokens, i.e. $\{\mathcal{H}^{j}\} \rightarrow \mathcal{Z} \times \mathcal{Z} \times \mathcal{Z}$, where $\mathcal{Z}$ is the vocabulary and three [MASK] tokens denoted as $z^1$, $z^2$, and $z^3$. Using the tailored prompt template featuring three [MASK]s and the verbalizer, the probability distribution over $\{\mathcal{H}^{j}\}$ can be formalized as the joint probabilities of $z^1$, $z^2$, and $z^3$,  i.e. $\text{Pr}(\mathcal{H}^j \mid \tilde{x}_i) = \text{Pr}(\mathcal{V}(\mathcal{H}^j) \mid \tilde{x}_i) = \text{Pr}(z_i^1 = h_3^j, z_i^2 = h_1^j, z_i^3 = h_2^j \mid \tilde{x}_i)$, where a hypothesis $\mathcal{H}^j$ consists of $h_1^j$ (the plausibility of $S_{i}^{1}$), $h_2^j$ (the plausibility of $S_{i}^{2}$), and $h_3^j$ (the probability of $H_{i}^{1}$ and $H_{i}^{2}$). Table~\ref{tab:Label_Set} summarizes the label words.
Given that T5 is able to predict masked tokens synchronously, the joint probability can be written as
\\
\scalebox{0.91}{\parbox{1.10\linewidth}{
\begin{align}
    \text{Pr}(\mathcal{H}^j \mid \tilde{x}_i) = \prod_{k=1}^{3}\text{Pr}(z_i^k = v^k(\mathcal{H}^j) \mid \tilde{x}_i), \label{eq:joint_prob}
\end{align}
}}
\\
where $v^k(\cdot): \{\mathcal{H}^{j}\} \rightarrow \mathcal{Z}$ is the submap of $\mathcal{V}(\cdot)$ for the $k$-th [MASK].
And then the final learning objective of $\alpha$-PACE is to maximize
\\
\scalebox{0.91}{\parbox{1.10\linewidth}{
\begin{align}
    \mathcal{J} = \frac{1}{|\mathcal{D}|} \sum_{(x_i, y_i) \in \mathcal{D}} \log \sum_{k=1}^{3} \text{Pr}(z_i^k = v^k(\mathcal{H}^j) \mid \tilde{x}_i).
\end{align}
}}
\\
The final prediction of $\mathcal{H}_i^{*}$ by choosing the maximum joint probability (i.e., self-consistency score) as Eq.~(\ref{eq:joint_prob}).

\subsection{General Self-Consistent Narrative framework}
The self-consistent framework introduced by previous studies may not be the optimal approach for the $\alpha$NLI task due to its lack of consideration for various narrative sequences~\cite{DBLP:journals/corr/abs-2203-11171}. Furthermore, ~\citet{bhagavatula2020abductive} and their follow-up works~\cite{du2021learning,paul2020social} only form probabilistic models focusing on the \textbf{Linear Chain Model} ($Pr(O^2|H^j)P(H^j|O^1)$, where $H^j$ can be $H^1$ or $H^2$) and \textbf{Fully Connected Model} ($Pr(O^2|H^j, O^1)P(H^j|O^1)$). This means their framework primarily considers the given fixed time sequence, i.e., $O^1$, $H^j$, and $O^2$, and may not align with the representation of the pre-trained language model. Therefore, we permute $\{O^{1}, O^{2}, (H^{1}, H^{2})\}$ and design six narrative sequence patterns for this task according to the orders of observations and the pair of hypotheses. The six patterns are illustrated in the overall framework in Figure~\ref{fig:framework}. For example, the O2HO1 sequence pattern means that we put $H^{j}$ in the middle of $O^2$ and $O^1$ and try to utilize the possible inter-sentential coherence information among them in this order.
After receiving the joint generation probabilities from each pattern, we normalize the probability distribution between ``$H^{1}$'' and ``$H^{2}$ '' to make it more contrastive. Then, we perform the majority voting over six narrative sequence pattern distributions and map token predictions to the label prediction.

\section{Experimental Setup}

\subsection{$\alpha$NLI Dataset}
The experiments are conducted on the ART dataset, aimed at assessing the performance of our model on the $\alpha$NLI task~\cite{bhagavatula2020abductive}. The observations of ART data were collected from a story corpus known as ROCstory~\cite{mostafazadeh-etal-2016-corpus}, while the corresponding hypotheses were generated by crowdsourcing. Moreover, $\alpha$NLI has a dedicated leaderboard with 3,040 test instances to measure the generalizability of the models. The detailed data statistics can be found in Table~\ref{tab:Dataset_Statistics}. By following previous work~\cite{bhagavatula2020abductive,du2021learning, zhu2020l2r2}, we employ accuracy as an evaluation metric to evaluate the empirical performance of our method in experiments and ablation studies.

\begin{table}[t]
\small
\centering
\begin{tabular}{c c c c}
\toprule
\textbf{Train} & \textbf{Dev} & \textbf{Test} & \textbf{Leaderboard} \\
\midrule
169,654 & 1,532 & 3,059 & 3,040 \\ 
\bottomrule
\end{tabular}
\vspace{-0.3 cm}
\caption{Statistics of ART and $\alpha$NLI leaderboard data.
}
\label{tab:Dataset_Statistics}
\vspace{-0.5 cm}
\end{table}

\subsection{Baselines}
The implementation detail of $\alpha$-PACE is displayed in Appendix~\ref{sec:Implementation}, and we compare $\alpha$-PACE with two categories of competitive baselines. The first category is the previous state-of-the-art (SOTA) baselines, such as \textit{BERT} \cite{devlin-etal-2019-bert}, \textit{RoBERTa} \cite{Liu2019RoBERTaAR}, \textit{DeBERTa} \cite{DBLP:conf/iclr/HeLGC21}, \textit{McQueen} \cite{Mitra2020HowAK}, \textit{MHKA} \cite{paul2020social}, \textit{ege-RoBERTa-large}  \cite{du2021learning}, \textit{$L2R^2$} \cite{zhu2020l2r2}, \textit{IMSL} \cite{li2021interactive}, \textit{UNIMO} \cite{DBLP:conf/acl/LiGNXLL0020}, and \textit{UNICORN (T5)}~\cite{DBLP:conf/aaai/LourieBBC21}. Another category of baselines is T5-based models, and we include the fine-tuned T5 model to illustrate the performance gain of our model. Furthermore, the prompt-based methods such as \textit{Prefix-Tuning (T5)}~\cite{DBLP:conf/acl/LiL20} and \textit{Prompt-Tuning (T5)}~\cite{DBLP:conf/emnlp/LesterAC21} are included as baselines for exhibiting the exact contribution of our proposed method.
The implementation details of the T5 fine-tuning model are described
in Appendix~\ref{sec: Appendix_t5_finetune_implementation}, while Prefix-Tuning and Prompt-Tuning methods are appended in Appendix~\ref{sec:Implementation_prefix_prompt}. More details of the baselines are in Appendix~\ref{sec:Appendix_baselines}.

\section{Experimental Results}

\begin{table}[!t]
\small
\centering
\scalebox{0.85}{
\begin{tabular}{lc c}
\toprule
\textbf{Model} & \textbf{Dev (\%)} & \textbf{Test (\%)}\\
\midrule
Random                                      & {-}     & {50.40}\\
BERT \cite{devlin-etal-2019-bert}     & {69.10} & {68.90}\\ 
RoBERTa \cite{Liu2019RoBERTaAR}       & {85.76} & {84.48}\\
McQueen \cite{Mitra2020HowAK}               & {86.68} & {-}\\
MHKA \cite{paul2020social}                  & {87.44} & {87.12}\\
ege-RoBERTa \cite{du2021learning}     & {-}     & {87.50}\\
$L2R^2$ \cite{zhu2020l2r2}                  & {88.44} & {86.11}\\
RoBERTa-L+IMSL \cite{li2021interactive}     & {89.20} & {-}\\

\hline
Prefix-Tuning (T5) \cite{DBLP:conf/emnlp/LesterAC21} & {84.20} & {83.88}\\
Prompt-Tuning (T5) \cite{DBLP:conf/acl/LiL20}& {86.23} & {85.98}\\
Fine-Tuning (T5) \cite{DBLP:journals/jmlr/RaffelSRLNMZLL20} & {87.07} & {87.68}\\
$\text{Fine-Tuning}_\text{General Consistency}$(T5)& {87.54} & {88.05}\\
\hline
$\alpha\text{-PACE}_\text{O1HO2 \& w/o Consistency}$(T5) & {90.60}  & {89.93}\\ 
$\alpha\text{-PACE}_\text{HO1O2 \& Task Consistency}$(T5) & 92.43 & 91.83\\
$\alpha\text{-PACE}_\text{General \& Task Consistency}$(T5)  & \bf{93.15} & \bf{92.54}\\
\hline
Human Performance & {-} & {91.40}\\
\bottomrule
\end{tabular}
}
\vspace{-0.3 cm}
\caption{The accuracy (\%) is evaluated on the $\alpha$NLI task. The approximation of learnable parameters for all models is displayed in Table~\ref{tab:tunable_parameters} in Appendix~\ref{sec:appendix_approximation_learnable parameters}. Fine-Tuning$_\text{General Consistency}$(T5) means the fine-tuned T5 model with our general narrative self-consistent framework, which considers six narrative sequences. $\alpha\text{-PACE}_\text{O1HO2 \& w/o Consistency }$ means this model with a single mask to predict the "choice1" or "choice2" and the O1HO2 pattern is the best model among all patterns.
}
\label{tab:Model_performance_fulldataset_1}
\vspace{-0.6 cm}
\end{table}

\subsection{Main Results}
Table~\ref{tab:Model_performance_fulldataset_1} and Table~\ref{tab:Model_performance_fulldataset_2} report the main experimental results on the $\alpha$NLI task, from which we derive the following conclusions. \textbf{First}, our model significantly outperforms all competitive baselines on the $\alpha$NLI task. Specifically, our method (i.e., $\alpha\text{-PACE}_\text{HO1O2}$) achieved a considerable improvement of 5.36\% on the development set, 4.15\% on the test set over the fine-tuning of the T5 model in the $\alpha$NLI task. It demonstrates that our model effectively utilizes a task-specific self-consistent method to validate the model's output and finalize a consistent answer. \textbf{Second}, $\alpha\text{-PACE}_\text{HO1O2}$ excels the prompt-based baselines (e.g., Prefix-Tuning and Prompt-Tuning) with at least 5.85\% test accuracy. This result exhibits the exact contribution of the task-specific self-consistent tailored prompt tuning-based model. \textbf{Third}, by adopting the general self-consistent narrative prompts, $\alpha\text{-PACE}_\text{General \& Task Consistency}$ obtains 92.54\% test accuracy and 92.01\% accuracy on the leaderboard test set. This result demonstrates that eliciting the inter-sentential coherence from the pre-trained language model and utilizing a general self-consistent framework considering six narrative sequences can partially solve this abductive reasoning task.

\begin{table}[!t]
\small
\centering
\scalebox{0.95}{
\begin{tabular}{l c }
\toprule
\textbf{Model} & \textbf{Leaderboard (\%)} \\
\midrule
Random                                & {50.41} \\
BERT \cite{devlin-etal-2019-bert}     &{66.75} \\ 
RoBERTa \cite{Liu2019RoBERTaAR}      &{83.91} \\ 
McQueen \cite{Mitra2020HowAK}        & {84.18} \\
ege-RoBERTa \cite{du2021learning}    & {85.95} \\
$L2R^2$ \cite{zhu2020l2r2}           & {86.81} \\
UNICORN (T5)\cite{DBLP:conf/aaai/LourieBBC21} & 87.34\\
RoBERTa-L+IMSL \cite{li2021interactive}   & {87.83} \\
DeBERTa \cite{DBLP:conf/iclr/HeLGC21}  & 89.70\\
DeBERTa(Ensemble) \cite{DBLP:conf/iclr/HeLGC21} & 90.00\\
UNIMO \cite{DBLP:conf/acl/LiGNXLL0020} & 91.18 \\ 
\hline
$\alpha\text{-PACE}_\text{O1HO2 \& w/o Consistency }$(T5)  & 89.51 \\ 
$\alpha\text{-PACE}_\text{HO1O2 \& Task Consistency}$(T5) & 91.61 \\ 
$\alpha\text{-PACE}_\text{General \& Task Consistency}$(T5)  & \bf{92.01}\\
\hline
Human Performance & {92.90}\\
\bottomrule
\end{tabular}}
\vspace{-0.3 cm}
\caption{The accuracy (\%) is evaluated on the test dataset from the $\alpha$NLI task leaderboard. The approximation of learnable parameters for all models is displayed in Table~\ref{tab:tunable_parameters} in Appendix~\ref{sec:appendix_approximation_learnable parameters}.
}
\label{tab:Model_performance_fulldataset_2}
\vspace{-0.4 cm}
\end{table}

\subsection{Ablation Study}
To better study the factors of the $\alpha\text{-PACE}$ model, we have devised numerous ablations on the joint probability for self-consistency, general narrative self-consistency, continuous prompt length, prompt engineering, and model size.

\paragraph{Joint Probability for Task-Specific Self-Consistency}
In our method, by estimating the likelihood of three masks to achieve self-consistency purposes, the dependencies of these three masks are exploited to enhance the ability of the pre-trained language model on this $\alpha\text{NLI}$ task. According to the experimental results in Table~\ref{tab:Joint_Probability_for_Self-Consistency}, we can conclude that
1) The performance of our task-specific self-consistent prompt model incorporating the signals from all three masks surpasses other models (e.g., $\alpha\text{-PACE}_\text{First \& Second}$), emphasizing the significance of dependencies and effectiveness of self-consistency;
2) The model with a single mask (e.g., $\alpha\text{-PACE}_\text{First}$), without integrating information from the other two masks, exhibits the worst performance;
3) The model with the third mask (e.g., $\alpha\text{-PACE}_\text{Second \& Third}$), which selects the best hypothesis, performs better than other models that lack the third mask. This finding highlights the importance and necessity of the third mask, summarizing the overall plausibility of two narrative sequences.

\begin{table}[!t]
\small
\centering
\begin{tabular}{l c c}
\toprule
\multicolumn{1}{c}{\textbf{Model}}
& \textbf{Dev (\%) } & \textbf{Test (\%)} \\
\hline
$\alpha\text{-PACE}_\text{First}$ & 87.41 & 86.94 \\
$\alpha\text{-PACE}_\text{Second}$ & 88.98 & 88.09 \\
$\alpha\text{-PACE}_\text{Third}$ & 90.60 & 89.93 \\
$\alpha\text{-PACE}_\text{First \& Second}$ & 88.78 & 89.03 \\
$\alpha\text{-PACE}_\text{First \& Third}$ & 90.24 & 89.48 \\
$\alpha\text{-PACE}_\text{Second \& Third}$ & 91.05 & 90.38 \\
\hline
\hline
$\alpha\text{-PACE}_\text{SM}$ & 91.20 & 90.13 \\
$\alpha\text{-PACE}_\text{General Consistency}$  & 91.78 & 90.64 \\
$\alpha\text{-PACE}_\text{SM \& Task Consistency}$ & 92.43 & 92.06 \\
$\alpha\text{-PACE}_\text{General \& Task Consistency}$  & \bf{93.15} & \bf{92.54} \\
\hline
\end{tabular}
\vspace{-0.2cm}
\caption{Ablation study in the joint probability for task-specific self-consistency and general narrative self-consistency on our model with HO1O2 patterns. $\alpha\text{-PACE}_\text{SM}$ means adopting the sample-and-marginalize method proposed by \citet{DBLP:journals/corr/abs-2203-11171} on our prompt model without the task-specific discrete prompt tokens.} 
\label{tab:Joint_Probability_for_Self-Consistency}
\vspace{-0.5 cm}
\end{table}

\paragraph{General Narrative Self-Consistency}
The prior research on the self-consistent prompt-based method (i.e., sample-and-marginalize method) relied on an individual prompt to sample various outputs and perform majority voting to resolve the inconsistency issue that language models suffered~\cite{DBLP:journals/corr/abs-2203-11171}. Therefore, we conducted experiments to compare the performance of both this method and our proposed general self-consistent method on our model, and the result is displayed in Table~\ref{tab:Joint_Probability_for_Self-Consistency}. By combining the likelihood of various outputs, the performance of sample-and-marginalize is slightly improving over the original single pattern-based model. Simultaneously, our general self-consistent approach surpasses this sample-and-marginalize method in two settings, with or without considering task consistency. Therefore, the results evidence the significance of our linguistic phenomenon-based self-consistent prompt, which considers various narrative sequences.

\paragraph{Prompt Length \& Prompt Engineering \& Model Size}
Furthermore, we conduct the ablation study on the continuous prompt length, prompt engineering, and the model size of our model in both few-shot and full training configurations. The details we described are in Appendix~\ref{sec: Appendix_ablation_study_pace}. The vital information worth mentioning is that without inserting prefix prompt and cloze prompt into our prompt template, the performance will significantly drop, and it illustrates the necessity of these two parts of learnable prompt tokens in our model.

\begin{table}[!t]
\small
\centering
\begin{tabular}{lc c c }
\toprule
\textbf{Model} & \textbf{Dev (\%)} & \textbf{Test (\%)}\\
\midrule
BERT-large  & 49.96\tiny{$\pm$0.73} & 50.79\tiny{$\pm$0.60}\\ 
RoBERTa-large & 58.20\tiny{$\pm$2.78} & 58.68\tiny{$\pm$2.80}\\
McQueen   & 60.38\tiny{$\pm$3.23} & 58.71\tiny{$\pm$2.25}\\
ege-RoBERTa-large & 65.80\tiny{$\pm$4.30} & 65.18\tiny{$\pm$3.27}\\ 
$L2R^2$ & 64.81\tiny{$\pm$2.40} & 65.68\tiny{$\pm$3.33}\\
Fine-Tuning(T5) & {69.57}\tiny{$\pm$2.01} & {71.18}\tiny{$\pm$2.06}\\
Prefix-Tuning(T5) & {73.96}\tiny{$\pm$5.36} & {72.29}\tiny{$\pm$5.10}\\
Prompt-Tuning(T5) & {76.34}\tiny{$\pm$1.70} & {75.38}\tiny{$\pm$1.83}\\
\hline
$\alpha\text{-PACE}_\text{O1HO2}$(T5) & 82.25\tiny{$\pm$1.09} & 81.90\tiny{$\pm$1.22} \\ 
$\alpha\text{-PACE}_\text{General Consistency}$(T5) &\bf{83.48}\tiny{$\pm$0.93} & \bf{83.15}\tiny{$\pm$0.93} \\
\bottomrule
\end{tabular}
\vspace{-0.2cm}
\caption{Model accuracy (\%) using 100 training instances compared with prompt-based models. We report the mean and standard deviation of five runs with different random seeds. }
\label{tab:valid_100instance}
\vspace{-0.3cm}
\end{table}

\begin{table}[!t]
\small
\centering
\begin{tabular}{l c}
\toprule
\multicolumn{1}{c}{\textbf{Model}}
& \textbf{Test (\%)} \\
\hline
Random & 50.40 \\
ChatGPT$_\text{Prompt}$ & 71.07 \\
ChatGPT$_\text{Task Consistency}$ & 72.57\\
ChatGPT$_\text{Sample-and-Marginalize}$  & 73.17\\
ChatGPT$_\text{General \& Task Consistency}$ & 74.42\\
\hline
\end{tabular}
\vspace{-0.2cm}
\caption{The performance of ChatGPT performs on the test set of $\alpha$NLI task. ChatGPT$_{Task Consistency}$ means utilizing the concatenate the $O^1$,$O^2$, and $H^j$ as the HO1O2 narrative patterns, instead of ChatGPT$_{Prompt}$ treat it as multi-choice questions. ChatGPT$_\text{Sample-and-Marginalize}$ means a prompt template sampling six times while the ChatGPT$_\text{General \& Task Consistency}$ means sampling with six narrative patterns.
}
\label{tab:ChatGPT_Performance}
\vspace{-0.7cm}
\end{table}

\begin{table*}[!ht]
\begin{minipage}{\textwidth}
    \begin{minipage}[b]{0.45\textwidth}
            \centering
            \includegraphics[width=\linewidth]{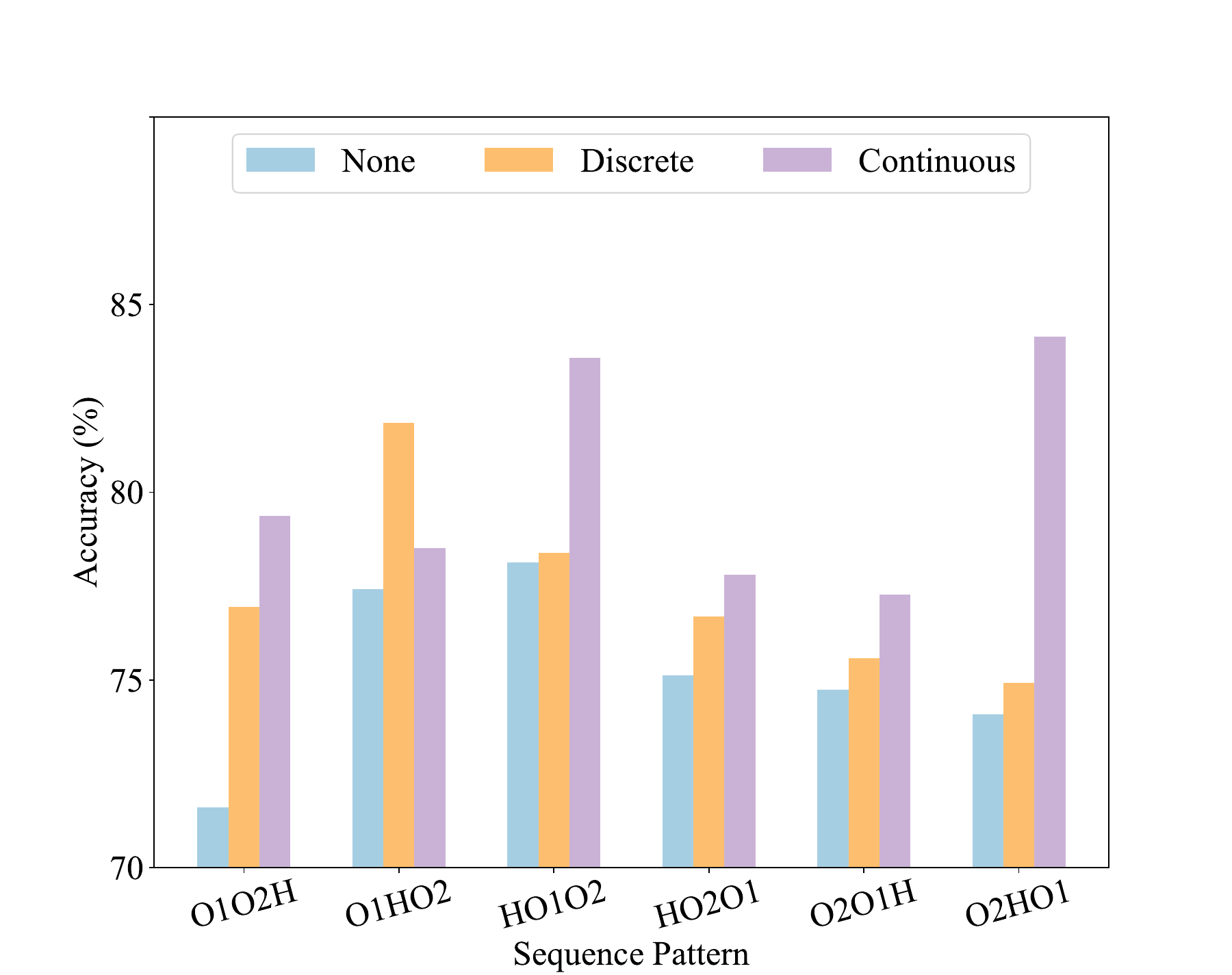}
            \vspace{-0.8 cm}
            \captionof{figure}{Performance comparison by adopting discourse connective in different settings. All models are training with 100 training instances.}
            \label{fig:Discourse_markers_performance}
            \vspace{-0.3 cm}
    \end{minipage}
    \hfill
    \begin{minipage}[b]{0.45\textwidth}
            \centering
            \begin{tabular}{l|c c}
            \toprule
            \textbf{Patterns} & \textbf{Discourse Connectives} \\
            \midrule
            O1O2H & meanwhile, in fact, because \\
            O1HO2 & in fact, as, as a result \\
            HO1O2 & because, meantime, as a result  \\ 
            HO2O1 & if, as a result, after \\ 
            O2O1H & meanwhile, if, as \\
            O2HO1 & in fact, as long as, because \\
            \bottomrule
            \end{tabular}
            \vspace{-0.2 cm}
            \caption{Top selected connectives for different patterns based on the model performance in the few-shot learning setting.}
            \label{tab:Discourse_markers_prompting}
            \vspace{-0.3 cm}
    \end{minipage}
\end{minipage}
\end{table*}

\begin{table*}[!ht]
    \centering
    \renewcommand\arraystretch{1.3}
    \scalebox{0.9}{
        \begin{tabular}{p{0.08\linewidth }|p{0.9\linewidth}}
            \toprule
            \textbf{Pattern} & \multicolumn{1}{c}{\textbf{Example}} \\
            \midrule
            O1O2H & \textbf{\textit{Meanwhile}}, \hlc[myorange]{Carl went to the store desperately searching for flour tortillas for a recipe.} \textbf{\textit{In fact}}, \hlc[myblue]{Carl left the store very frustrated} \textbf{\textit{because}} \hlc[mygreen]{the store had corn tortillas, but not flour ones.}\\
            \hline
            O1HO2 & \textbf{\textit{In fact}}, \hlc[myorange]{Carl went to the store desperately searching for flour tortillas for a recipe.} \textbf{\textit{As}} \hlc[mygreen]{the store had corn tortillas, but not flour ones.} \textbf{\textit{As a result}}, \hlc[myblue]{Carl left the store very frustrated.}\\
            \hline
            HO1O2 & \textbf{\textit{Because}} \hlc[mygreen]{the store had corn tortillas, but not flour ones.} \textbf{\textit{Meantime}}, \hlc[myorange]{Carl went to the store desperately searching for flour tortillas for a recipe.} \textbf{\textit{As a result}}, \hlc[myblue]{Carl left the store very frustrated.}\\ 
            \hline
            HO2O1 & \textbf{\textit{If}} \hlc[mygreen]{the store had corn tortillas, but not flour ones.} \textbf{\textit{As a result}}, \hlc[myblue]{Carl left the store very frustrated} \textbf{\textit{after}} \hlc[myorange]{Carl went to the store desperately searching for flour tortillas for a recipe.}\\ 
            \hline
            O2O1H & \textbf{\textit{Meanwhile}}, \hlc[myblue]{Carl left the store very frustrated} \textbf{\textit{if}} \hlc[myorange]{Carl went to the store desperately searching for flour tortillas for a recipe} \textbf{\textit{as}} \hlc[mygreen]{the store had corn tortillas, but not flour ones.}\\
            \hline
            O2HO1 & \textbf{\textit{In fact}}, \hlc[myblue]{Carl left the store very frustrated} \textbf{\textit{as long as}} \hlc[mygreen]{the store had corn tortillas, but not flour ones.} \textbf{\textit{Because}} \hlc[myorange]{Carl went to the store desperately searching for flour tortillas for a recipe.}\\ 
            \bottomrule
        \end{tabular}}
        \vspace{-0.2 cm}
        \captionof{table}{Case study for discourse connectives of different model patterns using the same case. The learned connectives are indicated in \textbf{boldface}.}
    \label{tab:Discourse_markers_case}
    \vspace{-0.5cm}
\end{table*}

\subsection{Few-Shot Setting}
\paragraph{Few-Shot Setting Comparing with Prompt-based methods}
With the sampled 100 training instances, we summarize our experimental results for the $\alpha$NLI task in Table~\ref{tab:valid_100instance}. We report the mean accuracy and standard deviation for five random seeds. As shown in Table~\ref{tab:valid_100instance}, our proposed model consistently outperforms other prompt-based models and appears more beneficial in this few-shot setting. Both our general consistency model and single narrative pattern model provide a significant gain over all stated baselines. 
With training on 100 instances, we observe that our proposed model with a single narrative sequence pattern significantly exceeds the Fine-Tuning (T5) in accuracy on the dev and test datasets by 12.68\% and 10.72\%, respectively. Furthermore, compared with the prompt-based models, our model still surpasses at least 6.52\% test accuracy. The large gap between our model and other T5-based models emphasizes the significance of the task-specific self-consistent method by considering the inter-sentential coherence information, proving that our model can effectively elicit and utilize temporal and causal information between observations and hypotheses. Moreover, after considering six narrative patterns, our $\alpha\text{-PACE}_\text{General Consistency}$ outperforms all our single pattern models by at least 1.23\% and 2.06\% in validation and test accuracy, respectively. We also study the influence of various training examples. We randomly subsample the entire dataset to obtain smaller datasets of size \{1, 5, 10, 20, 50\}. More details for the performance are shown in Figure~\ref{fig: Model performance different training instances} and Figure~\ref{fig: Our_Model_performance_different_training_instances} in Appendix~\ref{sec: fewer_Training_Instances}.

\subsection{Prompt Adaptation For ChatGPT}
With the powerful ability of LLMs exhibited on numerous tasks, we are curious about the capability of ChatGPT on zero-shot abductive commonsense reasoning tasks. 
Therefore, we test the ability of ChatGPT with four designed templates. The performance is shown in Table~\ref{tab:ChatGPT_Performance}. 
All the baselines can outperform random prediction.
ChatGPT$_{Task Consistency}$ improves the performance by 1.5\% over ChatGPT$_{Prompt}$ by utilizing the prompt template in the task-specific consistency method. We also find that the general self-consistent prompting (ChatGPT$_\text{General \& Task Consistency}$) demonstrates the additional performance boost over other baselines. Compared with ChatGPT$_\text{Sample-and-Marginal}$, instead of an individual prompt template sampling six times, our narrative framework, which considers six sequences, performs better on the $\alpha$NLI task.

\subsection{Interpretability}
An ideal interpretable prompt should be composed of natural language that makes it obvious why this prompt elicited such behavior from the model \cite{DBLP:conf/emnlp/LesterAC21}. Since the prompt tuning process only updates the prompt parameters and freezes the pre-trained language model, the learned prompt is expected to encode the inter-sentential coherence information (e.g., temporal and causal information) in our method. Therefore, the nearest neighbors discourse connectives of our learned cloze prompt (used to represent the discourse connectives in each pattern) should reasonably and appropriately describe the relationship between each sentence in various sentence sequences. The motivation of the interpretability section is to provide a view of the perspective of the significance of discourse connective and explore the possibility of different narrative sequences on the $\alpha$NLI task.

To obtain the nearest neighbors discourse connectives of these continuous cloze prompts in our method, we compute the cosine similarity between the averaged representation of learned cloze prompt tokens and the embedding vector of discourse connectives. The top selected connectives for each sequence pattern are shown in Table~\ref{tab:Discourse_markers_prompting}, and more details of discourse connectives can be found in Appendix~\ref{sec: Appendix_discourse_connectives_interpretability_section}. We use the data example utilized to illustrate the full-connect model in \citet{bhagavatula2020abductive} and insert the top selected connectives in between the sentences to form a narrative text, as shown in Table~\ref{tab:Discourse_markers_case}. We observe that the learned discourse connectives can describe the same collection of sentences in various sentence sequences in a rational and acceptable way. More case studies are shown in Table~\ref{tab:Discourse_markers_case_1} in Appendix~\ref{sec: Appendix_discourse_connectives_interpretability_section}.

We further test the performance on three input settings: (1) without continuous prompts inserted, (2) with inserting the top selected connectives as the discrete prompts, and (3) with the cloze continuous prompts. The results are shown in Figure~\ref{fig:Discourse_markers_performance}, and we see that the discrete connective is substantially superior to the without one. This finding underscores the plausibility and effectiveness of adopting the discourse marker to elicit coherent information from PLM. Moreover, the overall performance of our method with the continuous prompts outperforms the other two settings except for the O1HO2 pattern, where the discrete prompts are slightly better than the continuous prompts. It emphasizes the significance of utilizing continuous prompts to represent the connectives.

\section{Conclusion}
We developed a model that considers inter-sentential coherence and self-consistency through prompt tuning for improving the narrative understanding on the $\alpha$NLI task. Moreover, we propose a general self-consistent framework based on linguistic phenomena. The extensive experiments evidence the effectiveness of our proposed method.

\section*{Limitations}
Since all utilized information is only elicited from pre-trained language models (PLMs), our method relies on information or knowledge implicitly stored in the PLMs and the task dataset. This limitation restricts the capability owing to the reporting bias~\cite{DBLP:conf/cikm/GordonD13} in the pre-trained language models (PLMs). Moreover, our method is limited to the information type that can be elicited from PLMs. The future work for the constraint is to incorporate more abundant and sufficient knowledge to equip the model with more vital abilities. A possible method is adopting the grounding method~\cite{DBLP:conf/emnlp/LinCCR19} or retrieving the relevant nodes in the knowledge graph for each data instance, providing more contextual information and enhancing the capability of the model on this task.

\section*{Ethics Statement} \label{ethics_statement}
In this work, we conformed to accepted privacy practices and strictly followed the data usage policy. This paper presents a framework for guiding the PLM to understand the narrative context of input from the abductive natural language inference task. The ART dataset from the $\alpha$NLI task \cite{bhagavatula2020abductive} we used to train and evaluate the abductive inference ability of our model is publicly available, and this work is in the intended use. This dataset is collected from the manually curated story corpus ROCstory and should not contain any information that names or uniquely identifies individual people. Since we do not introduce social and ethical bias into the model or amplify any bias from the data, we can foresee no direct social consequences or ethical issues.

\section*{Acknowledgements}
The authors of this paper were supported by the NSFC Fund (U20B2053) from the NSFC of China, the RIF (R6020-19 and R6021-20) and the GRF (16211520 and 16205322) from RGC of Hong Kong. We also thank the support from NVIDIA AI Technology Center (NVAITC) and the UGC Research Matching Grants (RMGS20EG01-D, RMGS20CR11, RMGS20CR12, RMGS20EG19, RMGS20EG21, RMGS23CR05, RMGS23EG08).

\bibliography{anthology,custom}
\bibliographystyle{acl_natbib}

\clearpage

\appendix

\section{Appendix for Experimental Settings} \label{sec:appendix_1}
\subsection{Baselines} \label{sec:Appendix_baselines}
We compare $\alpha$-PACE with the following competitive baselines :\\
(a) \textit{BERT} \cite{devlin-etal-2019-bert} is a language model trained with masked-language modeling and the next sentence prediction objective.\\
(b) \textit{RoBERTa} \cite{Liu2019RoBERTaAR} is a powerful encoder that has the same architecture as BERT with robust optimization and more pre-training data.\\
(c) \textit{McQueen} \cite{Mitra2020HowAK} is a method to integrate external knowledge (e.g., commonsense knowledge) into a pre-trained language model (i.e., RoBERTa) to address the $\alpha$NLI task. \\
(d) \textit{MHKA} \cite{paul2020social} enhances RoBERTa by incorporating the social commonsense knowledge for the $\alpha$NLI task.\\
(e) \textit{ege-RoBERTa-large}  \cite{du2021learning} is a variational autoencoder-based model that learns commonsense knowledge by utilizing a latent variable for guiding the abductive reasoning task.\\
(f) \textit{$L2R^2$} \cite{zhu2020l2r2} reformulates the $\alpha$NLI task as a ranking problem using the learning-to-ranking framework to rank candidate hypotheses. \\   
(g) \textit{IMSL} \cite{li2021interactive} is an interactive language model that groups the correct/wrong hypotheses instead of ranking the hypotheses and adopts joint softmax focal loss for this $\alpha$NLI task.\\
(h) \textit{Fine-tuning (T5)}~\cite{DBLP:journals/jmlr/RaffelSRLNMZLL20} is an encoder-decoder model pre-trained on a multi-task mixture, where each task is converted into a text-to-text format. T5 performs well out of the box on many tasks by prepending a different prefix to the inputs.\\
(i) \textit{Prefix-Tuning (T5)}~\cite{DBLP:conf/acl/LiL20}: a method concatenates the tunable prefix tokens before the discrete input text, keeps language model parameters frozen, and optimizes these continuous task-specific prefix tokens. The implementation details of the Prefix-Tuning methods are appended in Appendix~\ref{sec:Implementation_prefix_prompt}.\\
(j) \textit{Prompt-Tuning (T5)}~\cite{DBLP:conf/emnlp/LesterAC21}: a vanilla Prompt Tuning-based model conditioning on a frozen model, releasing the constraints of the prompt templates from discrete to learnable prompts. The implementation details of the prompt tuning methods are appended in Appendix~\ref{sec:Implementation_prefix_prompt}.\\
(k) \textit{UNICORN (T5)}~\cite{DBLP:conf/aaai/LourieBBC21} is a universal commonsense reasoning model with multi-task pre-training based on T5-11b.\\
(l) \textit{DeBERTa}~\cite{DBLP:conf/iclr/HeLGC21} improves RoBERTa with disentangled attention and enhanced mask decoder training. It is only trained with half of the data used in RoBERTa.

\begin{figure*}[!ht]
\centering
\includegraphics[width=0.7\textwidth]{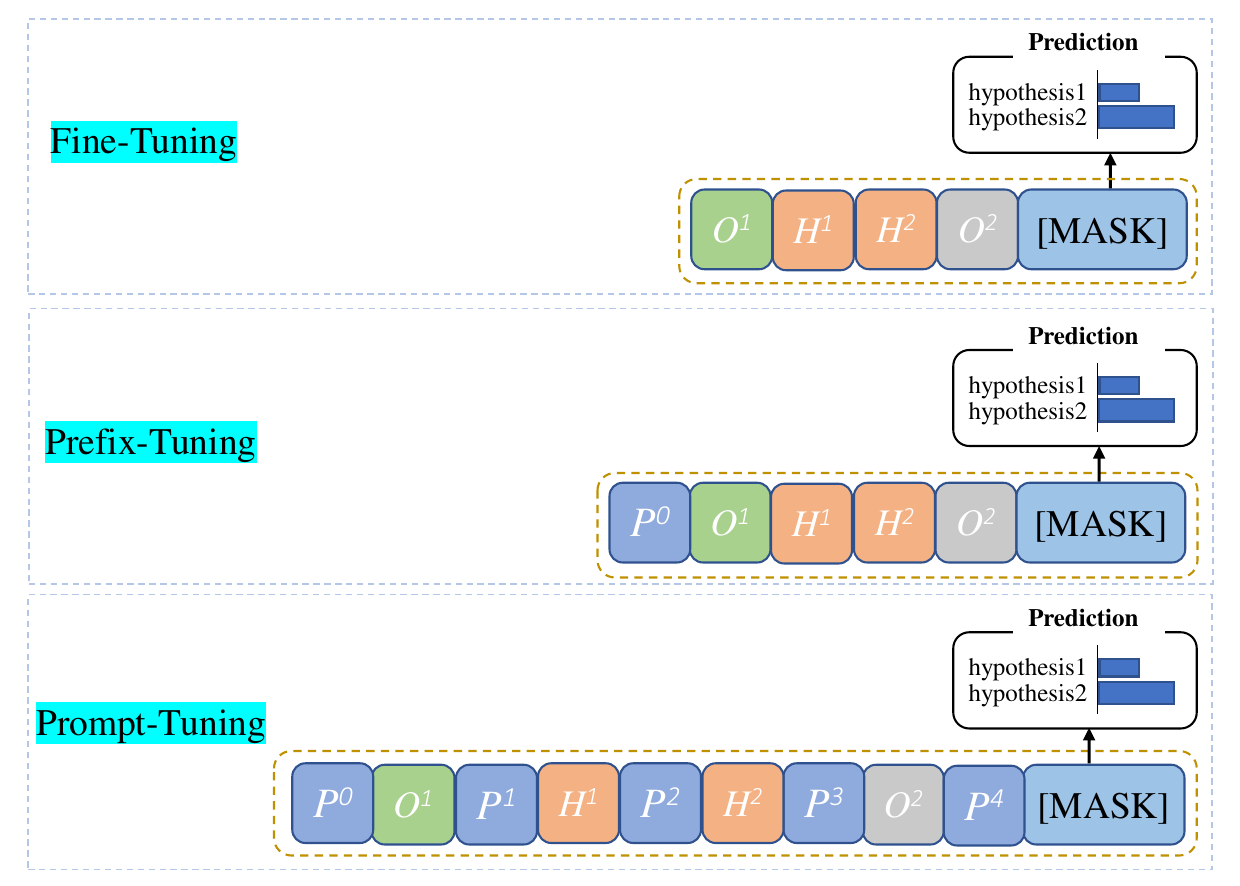}
\vspace{-0.1in}
\caption{
Fine-Tuning, Prefix-Tuning, and Prompt Tuning Templates. Two prompt tuning-based templates perform best among all designed templates in the template searching process for these baselines. The order of observations and hypotheses following the fully connected model proposed by~\citet{bhagavatula2020abductive}.
}
\label{fig:Prefix_Prompt_Tuning}
\vspace{-0.2in}
\end{figure*}

\subsection{$\alpha$-PACE Implementation Details}\label{sec:Implementation}
Our method is built upon the FLAN-T5~\cite{DBLP:journals/corr/abs-2210-11416} model was an enhanced version of the T5 model~\cite{DBLP:journals/jmlr/RaffelSRLNMZLL20} that has been finetuned in a mixture of tasks. We primarily use the 11B version but also experiment with various sizes (Small, Base, Large, and 3B versions) for the ablations. All the T5-based baselines are built upon the same FLAN-T5 model size (e.g., Fine-Tuning, Prompt-Tuning, and Prefix-Tuning). The general configuration follows the setting in \citet{DBLP:conf/emnlp/LesterAC21}. For the full-data training setting, the batch size and maximum sequence length are 1 and 350.
We set the prefix length $p^{0},p^{4}$ as 30, and all remaining cloze prompt lengths as 3.
We adopt an Adafactor optimizer by selecting a learning rate in \{8e-4, 8e-5, 6e-5, 5e-5, 3e-5\}, which yields the best performance on the dev set. The training is performed using cross-entropy loss, and the training steps are 30,000.

For the few-shot learning, we follow the full dataset setting except for the batch size and training steps being 3 and 5,000. Furthermore, we primarily use training set size K = 100 but explore K = \{1, 5, 10, 20, 50\} in the ablations. We sample the K examples from the full training data with five fixed seeds \{55, 58, 68, 72, 1,000\}. In this setting, we report the performance by averaging results along with the variance obtained for five different seeds.
Prompt tuning is conducted on two NVIDIA RTX A6000 GPUs, and it takes around 52 hours for full-data training and 3 hours for few-shot training.

\subsection{Implementation Details of T5 Model Fine-Tuning} \label{sec: Appendix_t5_finetune_implementation}
All the fine-tuning experiments are run on a server with 4 V100-32GB GPUs.
When fine-tuning the 11b version, we use DeepSpeed \cite{rajbhandari2020zero} with ZeRO stage 3 to offload parameters to memory.

\noindent \textbf{Model Input and Output}
The T5-based model serves as a competitive baseline in the main experiment by adopting the same model and model size.
We use the template \texttt{``Observation 1: \{\}$\backslash$nHypothesis 1: \{\}$\backslash$nHypothesis 2: \{\}$\backslash$nObservation 2: \{\}''} to transform a dataset instance into an input string.
The model is asked to generate either \texttt{Hypothesis 1} or \texttt{Hypothesis 2} as the predicted label. The order of two observations and two hypotheses following the fully connected model proposed by~\citet{bhagavatula2020abductive} and shown in Figure~\ref{fig:Prefix_Prompt_Tuning} in Appendix.

\noindent \textbf{Hyperparameter Search}
We first conduct a preliminary experiment to determine the range of hyper-parameters.
For base and large model sizes, we set the per-device train and validation batch size as 16 and 64, respectively.
For the 11b version, they are set as 8 and 32.
Then, we search for the optimal learning rate within \{3e-5, 1e-4, 3e-4\}.
The test performance of the model with the best validation accuracy is reported.

\subsection{Implementation Details of the Prefix-Tuning and Prompt Tuning}\label{sec:Implementation_prefix_prompt}

The prefix tuning~\cite{DBLP:conf/acl/LiL20} and prompt tuning~\cite{DBLP:conf/emnlp/LesterAC21} methods have been implemented as the baseline in full data learning and few-shot setting for comparison with our model. For a fair comparison, we count all the discrete textual tokens (non-tunable tokens) and the tunable tokens in our prompt template. There are 55 tokens, including 46 tunable tokens and nine textual tokens. In these two baselines, we will insert 55 tunable tokens into the respective prompt template. Moreover, we also adopt the same scale of T5 for these two baselines.

\paragraph{Prefix-Tuning} The overall configuration of this model follows the settings of prefix tuning~\cite{DBLP:conf/acl/LiL20}. The batch size and maximum sequence length of this model are 8 and 350. The training is performed using cross-entropy loss with an Adafactor optimizer~\cite{DBLP:conf/icml/ShazeerS18}. A learning rate selecting in \{3e-1, 5e-1, 8e-1\} yields the best performance on the validation set, and the training steps are 30,000. We insert 55 prefix tunable tokens into the prefix part of the input template. Since~\citet{bhagavatula2020abductive} stated that the given fixed time sequence (i.e., $O_1$, $H_i$, $O_2$) perform best among all the sequence, the order of two observations and two hypotheses is shown in Figure~\ref{fig:Prefix_Prompt_Tuning}.

\paragraph{Prompt-Tuning} The overall configuration of this model follows the settings of prompt tuning~\cite{DBLP:conf/emnlp/LesterAC21}. The batch size and maximum sequence length of this model are 8 and 350. The training is performed using cross-entropy loss, an Adafactor optimizer~\cite{DBLP:conf/icml/ShazeerS18}, and a learning rate selecting in \{3e-1, 5e-1, 8e-1\} yields the best performance on the validation set, with 30,000 training steps. We insert 55 tunable tokens evenly into inter-sentences or between sentences and mask tokens in the input template of this model. The order of two observations and two hypotheses is the same as the above method shown in Figure~\ref{fig:Prefix_Prompt_Tuning}.

\subsection{The Approximation of Learnable Parameters} \label{sec:appendix_approximation_learnable parameters}
To demonstrate the efficiency of our method, we attach the approximation of the learnable parameters for all models, including our model and the baselines. The approximation of the learnable parameters is displayed in Table~\ref{tab:tunable_parameters} in the Appendix. 

\begin{table}[!t]
\centering
\scalebox{0.9}{\begin{tabular}{l | c }
\hline
\textbf{Model} &  \textbf{Parameters}\\
\hline
BERT-large \cite{devlin-etal-2019-bert}           & 340M \\ 
RoBERTa-large \cite{Liu2019RoBERTaAR}    & 355M \\
McQueen \cite{Mitra2020HowAK}  & 355M \\
MHKA \cite{paul2020social}                       & 355M \\
ege-RoBERTa-large \cite{du2021learning}           & 355M \\
$L2R^2$  \cite{zhu2020l2r2}                  & 355M \\
IMSL  \cite{li2021interactive}              & 355M \\
DeBERTa-large \cite{DBLP:conf/iclr/HeLGC21} & 304M \\
UNICORN (T5)~\cite{DBLP:conf/aaai/LourieBBC21} & 11,000M \\
Prompt Tuning \cite{DBLP:conf/emnlp/LesterAC21}          & 1M \\ 
Prefix Tuning \cite{DBLP:conf/acl/LiL20}                 & 1M \\ 
Fine-Tuning \cite{DBLP:journals/jmlr/RaffelSRLNMZLL20}  & 11,000 M \\
\hline
$\alpha\textbf{-PACE}_\text{HO2O1}$                     & 34 M\\ 
\hline
\end{tabular}}
\vspace{-0.1cm}
\caption{The approximation of tunable parameters for models. Most baselines use RoBERTa-large as the backbone model, and their tunable parameters are approximated to be similar.}
\label{tab:tunable_parameters}
\vspace{-0.5cm}
\end{table}

\section{Appendix for Evaluation Result and Analysis}\label{sec:evaluation_result_analysis}


\subsection{Ablation study on the $\alpha$-PACE} \label{sec: Appendix_ablation_study_pace}

\paragraph{Prompt Length} 
\begin{figure}[t]
    \centering
    \includegraphics[width=\linewidth]{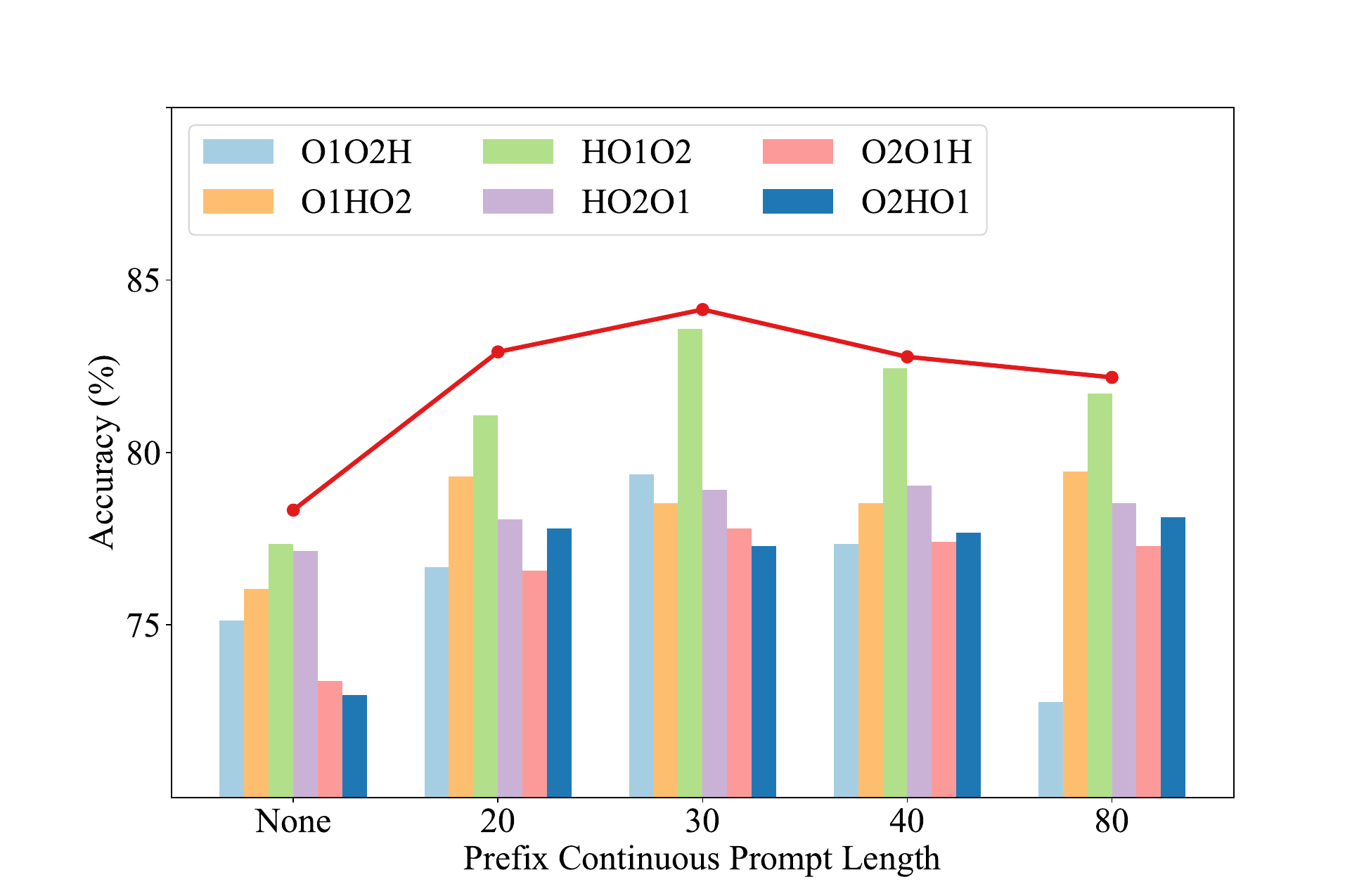}
    \caption{Performance of our method with different numbers of prefix continuous prompt tokens ($p^0,p^4$) on the test dataset using 100 training instances. 
    The red line indicates the performance of $\alpha\text{-PACE}_\text{General \& Task Consistency}$.
    }
    \label{tab:prefix continuous prompt}
    \vspace{-0.5cm}
\end{figure}

\begin{figure}[t]
    \includegraphics[width=\linewidth]{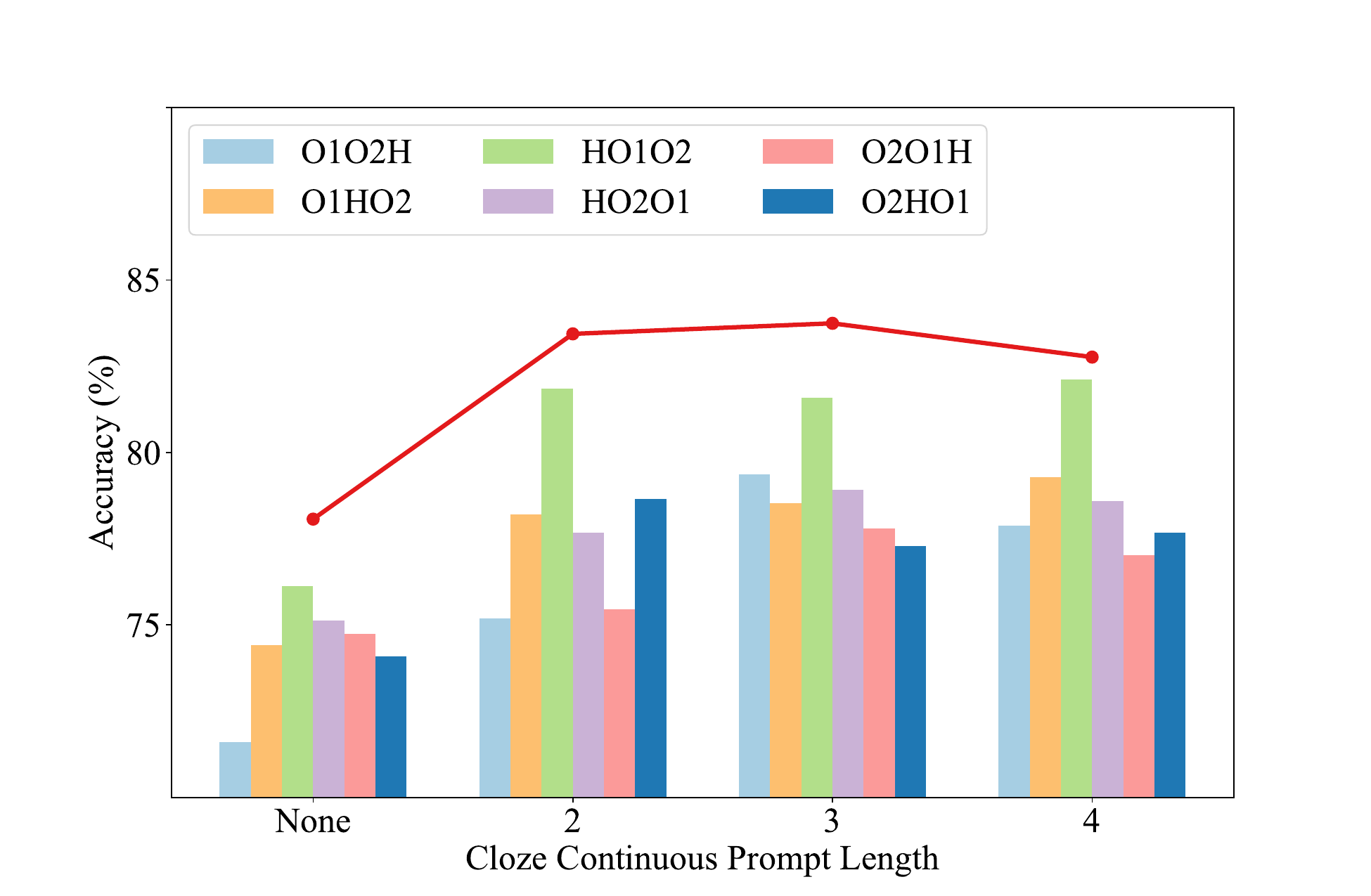}
    \caption{Performance of our method with different numbers of cloze continuous prompt tokens ($p^1, p^2, p^3, p^5, p^6, p^7$) on the test dataset using 100 training instances. The red line indicates the performance of $\alpha\text{-PACE}_\text{General \& Task Consistency}$.
    }
    \label{fig:connective continuous prompt}
    \vspace{-0.5cm}
\end{figure}

Within our designed prompt template, two parts of continuous prompts are concatenated with the input sentences.
The first part is two prefix prompts with $p^0$ and $p^4$ tokens inserted before template tokens ``choice1'' and ``choice2''.
The other part is the cloze prompts inserted to two positions: $p^1$ (or $p^5$) tokens between ``choice1'' (or ``choice2'') and the first input sentence, and $p^2, p^3$ (or $p^6, p^7$) tokens between input sentences (observations or hypotheses).

For the prefix prompt, we train prompts for our model on 100 training instances by varying the prefix prompt length in \{None, 20, 30, 40, 80\} while keeping the other setting unchanged. Figure \ref{tab:prefix continuous prompt} shows that the performance of most models with continuous prefix prompts exceeds the ``None'' one. Inserting the prefix prompt is critical to achieving good performance. After increasing beyond 30 prefix prompt tokens, the performance for different patterns becomes unstable, and some patterns yield low performance, which hurts the performance of the voting method.

For the cloze prompt, we tune our model on 100 training instances by varying the cloze prompt length in \{None, 2, 3, 4\} while fixing other settings. The result is given in Figure \ref{fig:connective continuous prompt}, and the overall performance of our model with the cloze prompt is better than the ``None'' one. Hence, inserting the cloze prompt is another essential factor in obtaining good performance. With the cloze prompt excess of three prompt tokens, the performance of each pattern does not improve significantly, and the performance of the voting method falls.

\begin{figure}[!t]
    \centering
    \includegraphics[width=\linewidth]{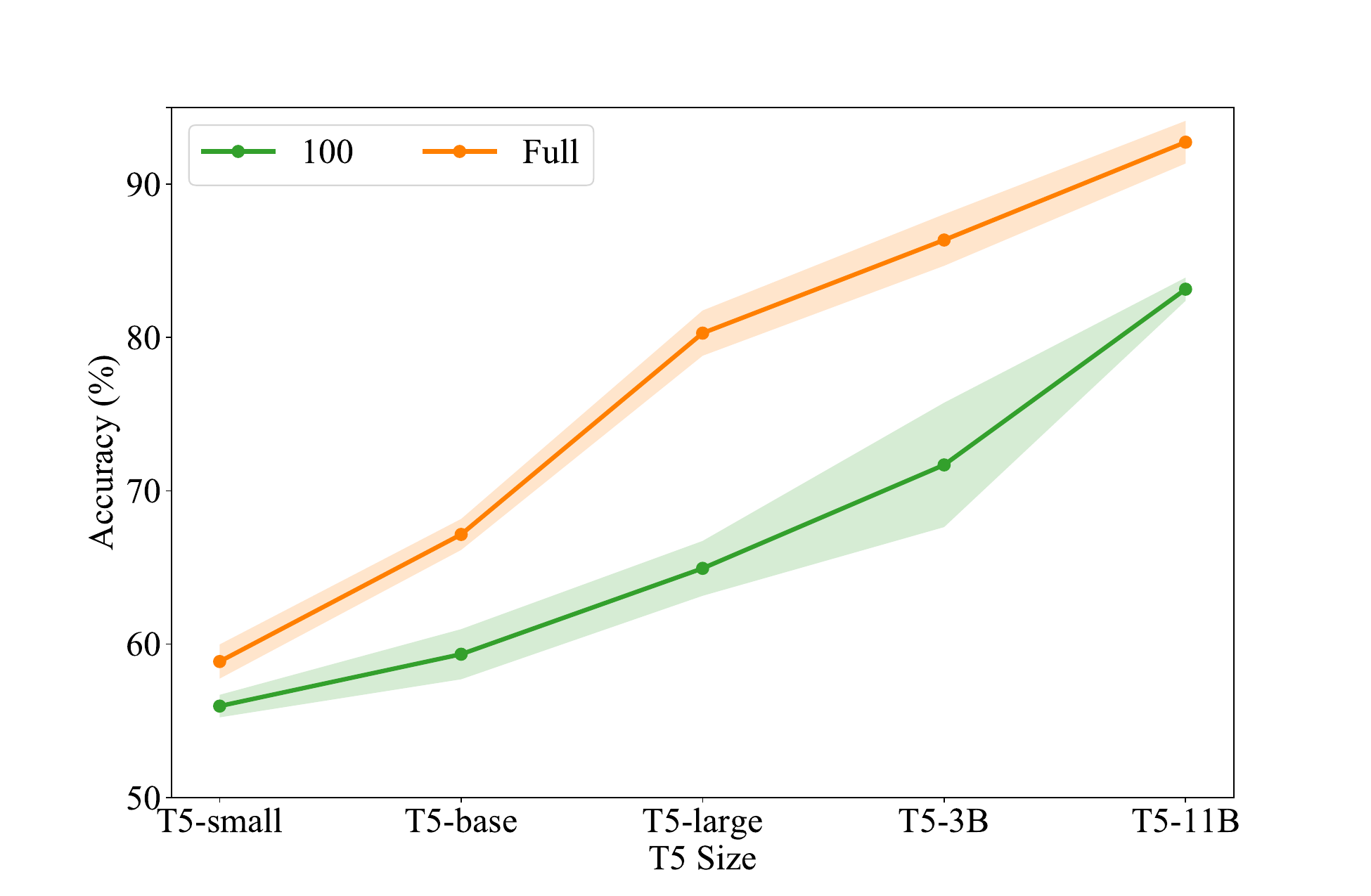}
    \vspace{-0.8 cm}
    \caption{Performance (test accuracy \%) comparison on various T5 model sizes in the few-shot and full training settings on the test set.}
    \label{fig:Model Size}
    \vspace{-0.3 cm}
\end{figure}

\paragraph{Prompt Engineering}
\begin{figure*}[t]
    \centering  \includegraphics[width=\linewidth]{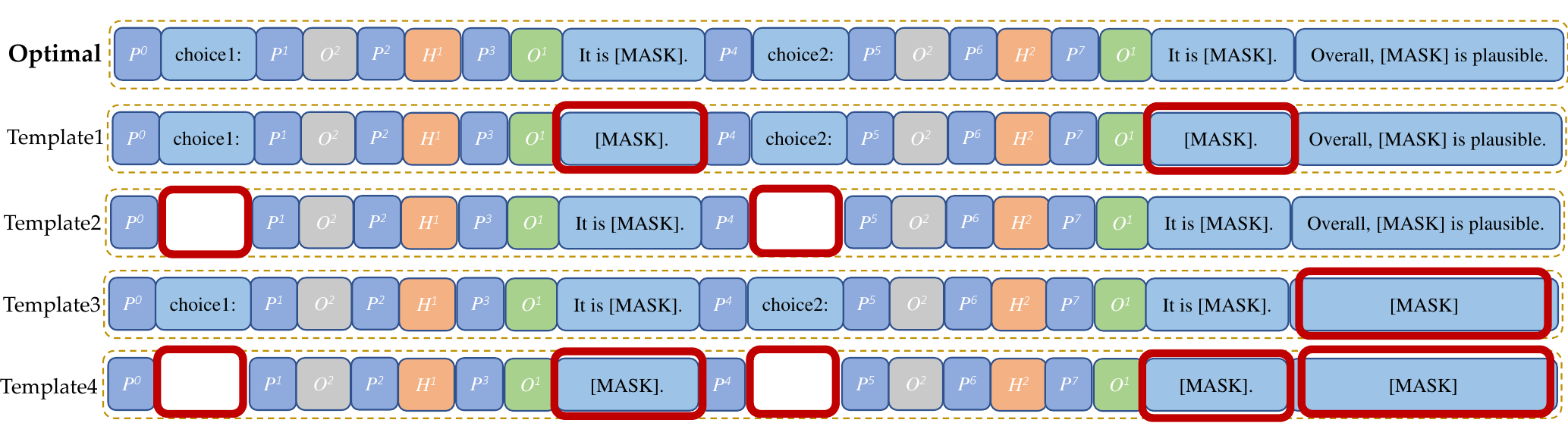}
    \vspace{-0.8 cm}
    \caption{$\alpha\text{-PACE}$ prompt template searching. The ``Optimal Templates'' is the finalized optimal template for implementing experiments to compare with extensive baselines. The red rectangle highlights the modified parts.}
    \label{fig:DiscoPrompt_Template_Searching}
    \vspace{-0.5 cm}
\end{figure*}

\begin{table}[!ht]
\centering
\begin{tabular}{l|c|c}
\hline
\textbf{Models}
& \textbf{Dev} & \textbf{Test} \\
\hline
\textbf{Optimal Template} & \textbf{82.25\tiny{$\pm$1.09}}& \textbf{81.90\tiny{$\pm$1.22}} \\ 
Template 1 & 81.81\tiny{$\pm$0.59} & 80.69\tiny{$\pm$1.19} \\ 
Template 2 & 81.78\tiny{$\pm$0.34} & 81.59\tiny{$\pm$1.90} \\
Template 3 & 81.98\tiny{$\pm$1.09} & 81.26\tiny{$\pm$1.22} \\
Template 4 & 80.61\tiny{$\pm$0.91} & 80.48\tiny{$\pm$0.95} \\
\hline
\end{tabular}
\caption{
Ablation study on the discrete textual prompt tokens of $\alpha\text{-PACE}$ on $\alpha$NLI task in few-shot (100 instances) setting.
}
\label{table:discrete_token_methods_ablation}
\vspace{-0.6cm}
\end{table}

Apart from the continuous prompt in our designed prompt template, there are some tokens in natural textual form and discrete non-tunable tokens. As shown in Figure~\ref{fig:DiscoPrompt_Template_Searching}, we gradually remove different portions of these discrete textual tokens and evaluate their importance. The performance shown in Table~\ref{table:discrete_token_methods_ablation} demonstrates that all discrete textual prompts are essential for achieving satisfactory performance compared with those without manual tips (i.e., Template 4). Among all discrete prompt tokens, the portion "It is <mask>" significantly affects the performance of our model as this discrete part facilitates eliciting the evaluation of PLMs on the plausibility of two hypotheses.

\paragraph{Model Size}
We compare the performance of various T5 model sizes in both few-shot and full training configurations. As demonstrated in Figure~\ref{fig:Model Size}, increasing the model size from T5-Large to T5-11B results in an average accuracy increase of around 10\% for the few-shot setting and 6\% for the full data learning setting. The results imply that a larger model encodes richer narrative knowledge and is advantageous for effectively eliciting more narrative knowledge, which is also our motivation for experimenting with pre-trained models as large as feasible. 

\subsection{Training Instances} \label{sec: fewer_Training_Instances}
\begin{figure*}[!t]
\begin{minipage}{\textwidth}
    \begin{minipage}[b]{0.48\textwidth}
        \centering
        \includegraphics[width=1.05\linewidth]{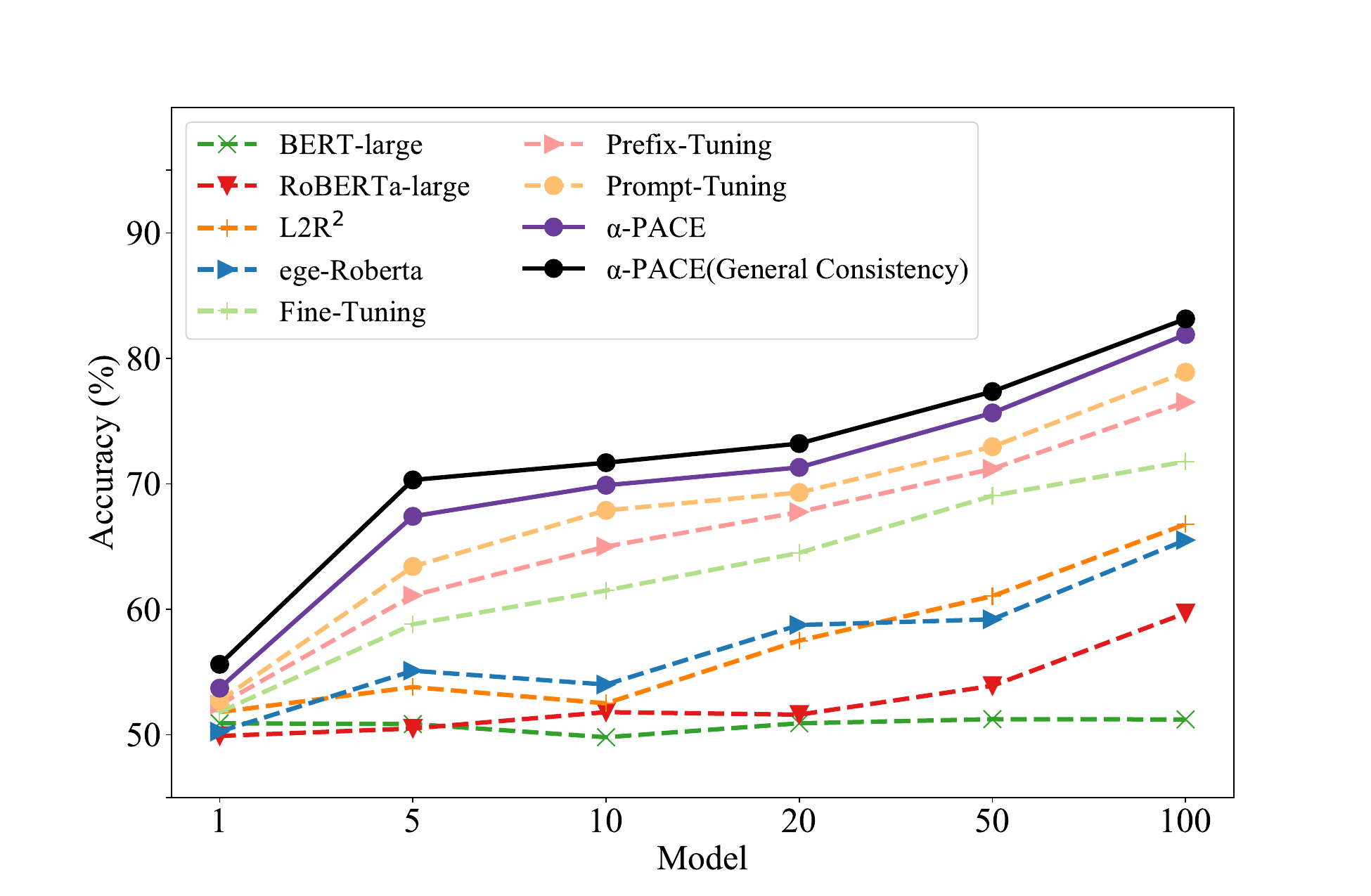}
        \vspace{-0.7cm}
        \caption{Model performance comparison by using various numbers of the training instances on the test set. 
        }
        \label{fig: Model performance different training instances}
    \end{minipage}
    \hfill
    \begin{minipage}[b]{0.48\textwidth}
        \centering
        \includegraphics[width=1.05\linewidth]{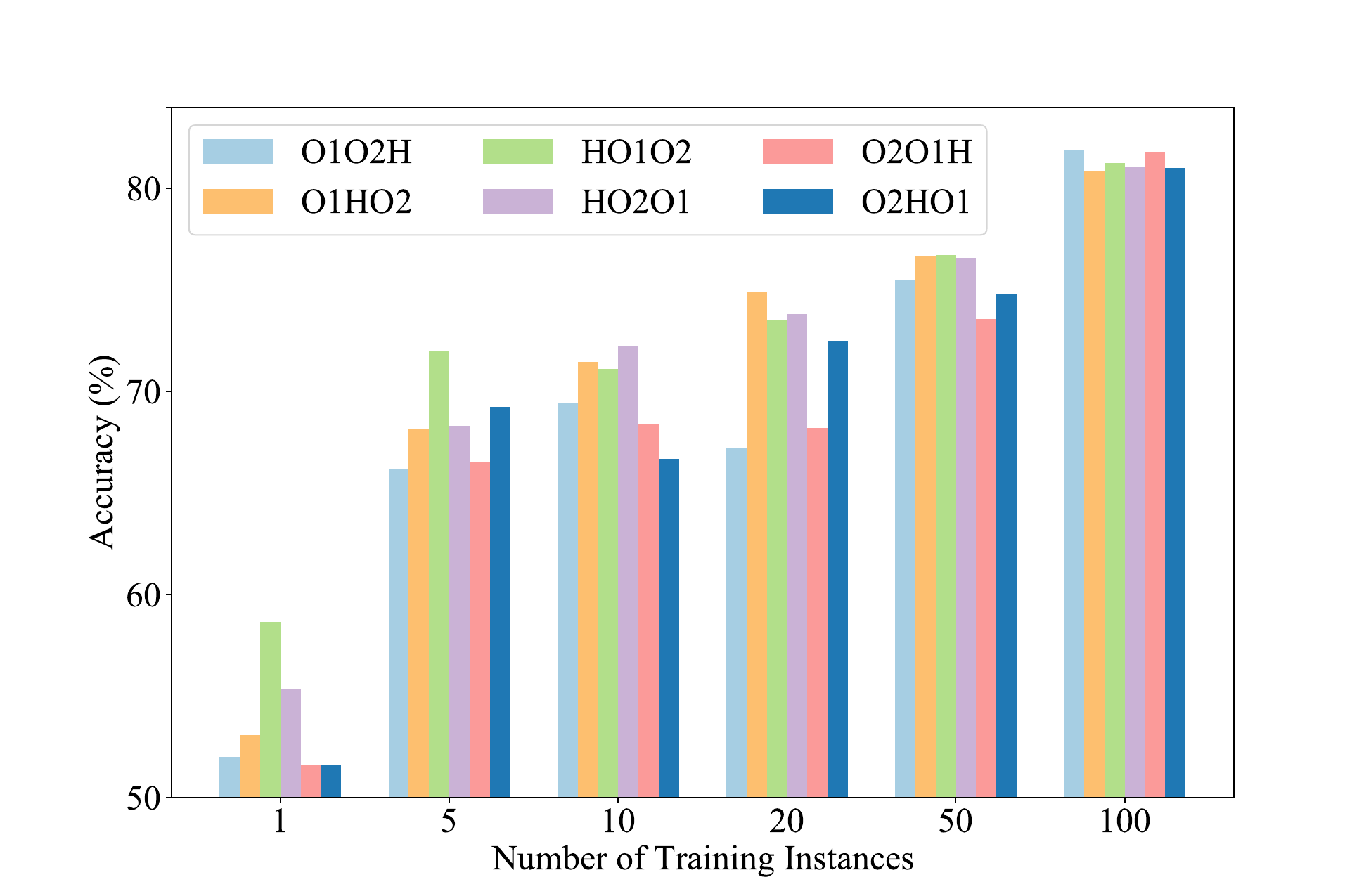}
        \vspace{-0.7cm}
        \captionof{figure}{Model performance comparison by using various numbers of the training instances on the test set.}
        \label{fig: Our_Model_performance_different_training_instances}    
    \end{minipage}
    \vspace{-0.5 cm}
\end{minipage}

\end{figure*}

To further study the influence of various training examples. We randomly subsample the entire dataset to obtain smaller datasets of size \{1, 5, 10, 20, 50\}. More training examples means more narrative context information our model can learn. Figure~\ref{fig: Model performance different training instances} shows that the average performance of six patterns of $\alpha\text{-PACE}$ increases as the number of training instances increases and consistently keeps a large gap against other baselines (e.g., RoBERTa). Interestingly, with a single training instance, the performance of the HO1O2 pattern (from Figure~\ref{fig: Our_Model_performance_different_training_instances}) achieves 57.37\% test accuracy, much greater than the other five patterns in our model. Therefore, employing an instance-specific narrative sequence pattern may give the pre-trained model a better outcome on limited training instances. This again verifies the motivation for why we involve six narrative sequence patterns in our method. 
Furthermore, all prompt-based methods received excellent performance in this few-shot setting, consistent with previous works~\cite{DBLP:conf/emnlp/LesterAC21, DBLP:conf/acl/LiL20}. Furthermore, when baselines train with ten times more data than our method, our single model still outperforms most of these baselines. For example, in Figure~\ref{fig: Model performance different training instances}, the performance of $\alpha\textbf{-PACE}_\text{O1HO2}$ training with five instances significantly exceeds almost all baseline training with 50 instances.

\subsection{Discourse Connectives in Interpretability Section} \label{sec: Appendix_discourse_connectives_interpretability_section}
The Penn Discourse Treebank 2.0 (PDTB 2.0) is a commonly used dataset in discourse parsing tasks and is a large-scale corpus containing 2,312 Wall Street Journal (WSJ) articles annotated by experts~\cite{prasad-etal-2008-penn}. There are many discourse connectives in PDTB 2.0 that belong to various discourse relations. Discourse relations are of utmost importance for achieving textual coherence and are deemed an essential step for a multitude of downstream tasks that involve more context, including but not limited to question answering~\cite{DBLP:conf/acl/JansenSC14}, text generation~\cite{DBLP:conf/naacl/BosselutCHGHC18}, and argument mining~\cite{DBLP:conf/acl/LiuOSJ20, 10.1145/3587716.3587743}.

To obtain the nearest neighbors discourse connectives of these continuous cloze prompts in our method, we compute the cosine similarity between the averaged representation of learned cloze prompt tokens and the embedding vector of discourse connectives. 
We acquired these discourse connectives from the Penn Discourse Treebank 2.0~\cite{prasad-etal-2008-penn}, a commonly used dataset in discourse analysis. These connectives are composed of two main categories of discourse relations: \textbf{\textit{Contingency}} and \textbf{\textit{Temporal}}. There are 23 connectives in these two categories after removing duplicates. The details of connectives can be found in Table~\ref{tab:PDTB2.0} and Figure~\ref{tab:PDTB2.0_connctives} in Appendix. The top selected connectives for each sequence pattern are shown in Table~\ref{tab:Discourse_markers_prompting} in the few-shot setting. We use the data example utilized to illustrate the full-connect model in \citet{bhagavatula2020abductive} and insert the top selected connectives in between the sentences to form a narrative text, as shown in Table~\ref{tab:Discourse_markers_case}. We observe that the learned discourse connectives can describe the same collection of sentences in various sentence sequences in a rational and acceptable way. More case studies are shown in Table~\ref{tab:Discourse_markers_case_1} in Appendix~\ref{sec: Appendix_discourse_connectives_interpretability_section}.

\subsection{ChatGPT Capability on Abductive Commonsense Reasoning}\label{sec:appendix_ChatGPT_Capability_on_Abductive_Commonsense_Reasoning}
The impressive ability of instruction-following large language models (e.g.,  ChatGPT~\cite{openai2022chatgpt} and GPT-4~\cite{DBLP:journals/corr/abs-2303-08774}) has been exhibited by many studies~\cite{DBLP:journals/corr/abs-2303-12712,
DBLP:journals/corr/abs-2302-04023, 
DBLP:journals/corr/abs-2302-10724,
DBLP:journals/corr/abs-2304-14827,
alpaca, 
vicuna2023, 
DBLP:journals/corr/abs-2305-12870}. There are some challenges remain unresolved such as the associated ethical implications and privacy concerns~\cite{DBLP:journals/corr/abs-2304-05197, DBLP:journals/corr/abs-2212-09292, DBLP:journals/corr/abs-2302-00539}. In this work, we are curious about the capability of ChatGPT on zero-shot abductive commonsense reasoning tasks. Hence, we test the ability of ChatGPT~\footnote{The evaluation is performed in February 2023 by calling ChatGPT API.} with four designed templates on the test set of $\alpha$NLI task. 
ChatGPT$_{prompt}$ reformulate the task as the multi-choice questions to predict the class label by following~\citet{DBLP:journals/corr/abs-2302-04023}.
ChatGPT$_{Task Consistency}$ concatenates the $O^1$,$O^2$, and $H^j$ as two narrative sentence sequences, which is the same as the prompt template shown in Figure~\ref{fig:framework}.
ChatGPT$_\text{Sample-and-Marginalize}$ is to sample six times with an individual prompt template.
ChatGPT$_\text{General \& Task Consistency}$ utilizes six narrative patterns as the input template and each time only feeds only one narrative pattern. Furthermore, it is imperative to note that the input template, which incorporates in-context learning, is heavily dependent on the chosen training examples that form the prefix demonstration of the prompt template. The performance of in-context learning is subject to high variance based on the specific examples chosen, the quantity of examples, as well as the order in which they are presented. Consequently, this particular template has been excluded from consideration in this particular section. The performance of the random guess model is derived via the averaging of the results obtained from five distinct iterations.

\onecolumn

\FloatBarrier

\begin{table*}[!ht]
    
    \begin{minipage}{\textwidth}
        \begin{minipage}[b]{0.68\textwidth}
            \centering
            \scriptsize
            \begin{tabular}{l l}
                \toprule
                \textbf{Relation} & \multicolumn{1}{c}{\textbf{Connectives}} \\
                \midrule
                \textbf{Contingency} & when (4), furthermore (1), indeed (1) \\
                \textbf{Contingency.Cause} & as (1) \\
                \textbf{Contingency.Cause.Reason} & because (2098), as (502), since (248) \\
                \textbf{Contingency.Cause.Result} & so (843), as a result (271), thus (220) \\
                \textbf{Contingency.Condition.Factual past} & if (5) \\
                \textbf{Contingency.Condition.Factual present} & if (48), if then (7), when (7) \\
                \textbf{Contingency.Condition.General} & if (145), when (124), as long as (6) \\
                \textbf{Contingency.Condition.Hypothetical} & if (563), when (22), if then (20) \\
                \textbf{Contingency.Condition.Unreal past} & if (47), if then (1) \\
                \textbf{Contingency.Condition.Unreal present} & if (87) \\
                \textbf{Contingency.Pragmatic cause.Justification} & because (30), as (13), in fact (7) \\
                \textbf{Contingency.Pragmatic condition.Implicit assertion} & if (23), when (11), because (2) \\
                \textbf{Contingency.Pragmatic condition.Relevance} & if (12), when (1), so (1) \\
                \hline
                \textbf{Temporal} & when (6), meanwhile (1), while (1) \\
                \textbf{Temporal.Asynchronous} & before and after (1), meantime (1), in turn (1) \\
                \textbf{Temporal.Asynchronous.Precedence} & then (556), before (240), until (100) \\
                \textbf{Temporal.Asynchronous.Succession} & after (452), when (181), previously (124) \\
                \textbf{Temporal.Synchrony} & when (475), as (427), while (236) \\
                \bottomrule
            \end{tabular}
            \caption{Penn Discourse Treebank 2.0 contingency and temporal connectives, where frequencies are reported in brackets.}
            \label{tab:PDTB2.0}
        \end{minipage}
        \hfill
        \begin{minipage}[b]{0.3\textwidth}
            \centering
            \includegraphics[width=\linewidth]{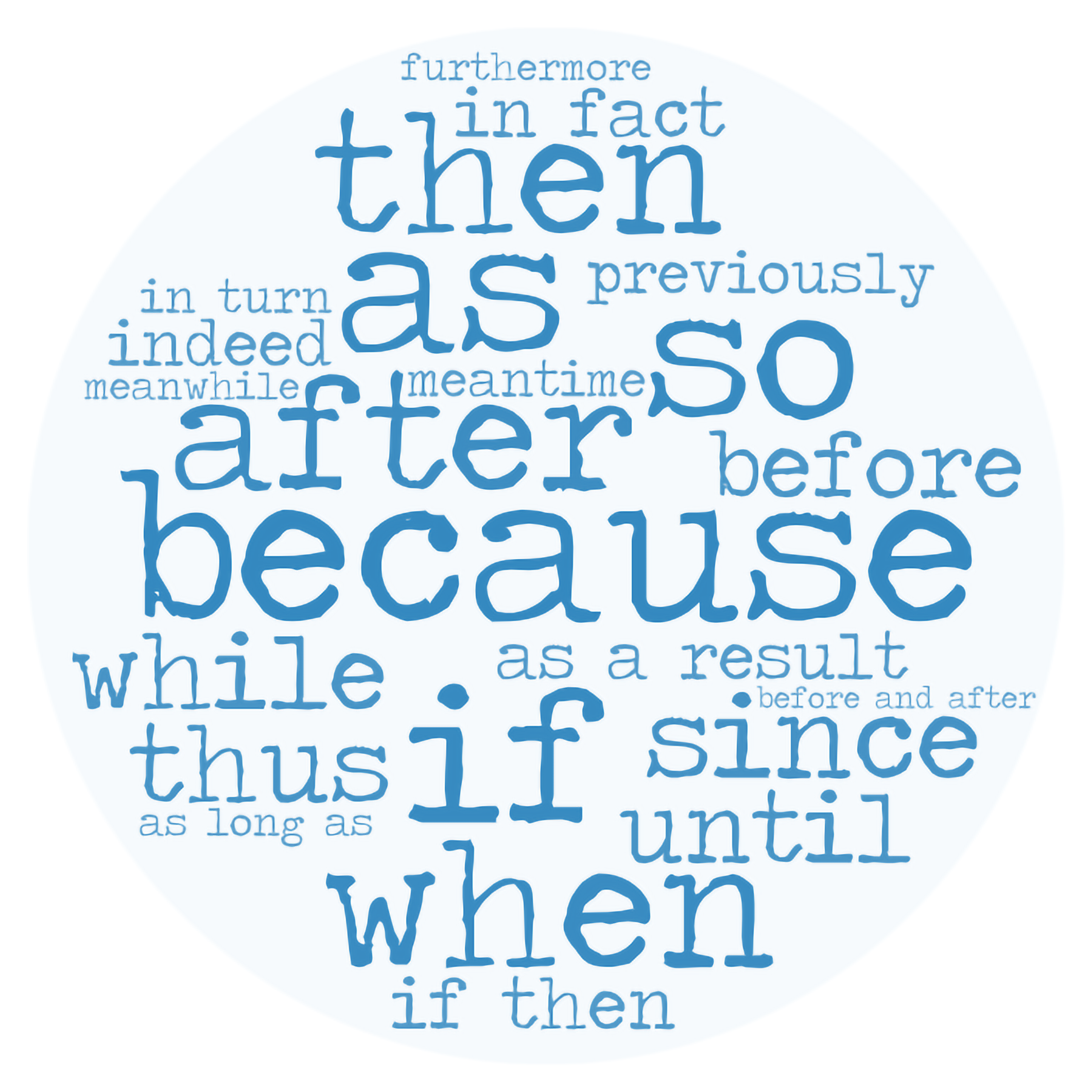}
            \captionof{figure}{Penn Discourse Treebank 2.0 top frequency discourse markers.}
            \label{tab:PDTB2.0_connctives}
        \end{minipage}
    \end{minipage}
\end{table*}

\FloatBarrier

\begin{table*}[!ht]
    \centering
    \scalebox{0.95}{
    \begin{tabular}{p{0.08\linewidth}|p{0.9\linewidth}}
        \toprule
        \textbf{Pattern} & \multicolumn{1}{c}{\textbf{Example}} \\
        \midrule
        O1O2H & \textbf{Meanwhile}, Jimmy had to get a root canal. \textbf{In fact}, He did not feel a thing and the procedure went smoothly \textbf{because} Jimmy got plenty of novocaine for the procedure. \\
        
        O1HO2 & \textbf{In fact}, Jimmy had to get a root canal.\textbf{As} Jimmy got plenty of novocaine for the procedure. \textbf{As a result}, He did not feel a thing and the procedure went smoothly. \\
        
        HO1O2 & \textbf{Because} Jimmy got plenty of novocaine for the procedure. \textbf{Meantime}, Jimmy had to get a root canal.  \textbf{As a result}, he did not feel a thing and the procedure went smoothly.  \\ 
        
        HO2O1 & \textbf{If} Jimmy got plenty of novocaine for the procedure. \textbf{As a result}, he did not feel a thing and the procedure went smoothly \textbf{after} Jimmy had to get a root canal.\\ 
        
        O2O1H & \textbf{Meanwhile}, he did not feel a thing and the procedure went smoothly \textbf{if} Jimmy had to get a root canal \textbf{as} Jimmy got plenty of novocaine for the procedure.\\
        
        O2HO1 & \textbf{In fact}, he did not feel a thing and the procedure went smoothly \textbf{as long as} Jimmy got plenty of novocaine for the procedure. \textbf{Because} Jimmy had to get a root canal.\\ 
        \hline

        O1O2H & \textbf{Meanwhile}, Jane was a professor teaching piano to students. \textbf{In fact}, Jane spent the morning sipping coffee and reading a book \textbf{because} none of Jane's students had a lesson that day. \\
        
        O1HO2 & \textbf{In fact}, Jane was a professor teaching piano to students. \textbf{As} none of Jane's students had a lesson that day. \textbf{As a result}, Jane spent the morning sipping coffee and reading a book. \\
        
        HO1O2 & \textbf{Because} none of Jane's students had a lesson that day. \textbf{Meantime}, Jane was a professor teaching piano to students. \textbf{As a result}, Jane spent the morning sipping coffee and reading a book. \\ 
        
        HO2O1 & \textbf{If} none of Jane's students had a lesson that day. \textbf{As a result}, Jane spent the morning sipping coffee and reading a book \textbf{after} Jane was a professor teaching piano to students.\\ 
        
        O2O1H & \textbf{Meanwhile}, Jane spent the morning sipping coffee and reading a book \textbf{if} Jane was a professor teaching piano to students \textbf{as} none of Jane's students had a lesson that day.\\
        
        O2HO1 & \textbf{In fact}, Jane spent the morning sipping coffee and reading a book \textbf{as long as} none of Jane's students had a lesson that day \textbf{because} Jane was a professor teaching piano to students.\\ 
        \bottomrule
    \end{tabular}
    }
    \caption{Case study for the discourse connectives of different model patterns using the same data example. The learned connectives are highlighted in \textbf{bold}.}
    \label{tab:Discourse_markers_case_1}
\end{table*}

\end{document}